\documentclass[letterpaper, 10 pt, conference]{ieeeconf}

\IEEEoverridecommandlockouts
\usepackage{graphicx}
\usepackage{amsmath}
\usepackage{amssymb}

\usepackage{booktabs}
\usepackage{multirow}
\usepackage[table]{xcolor}
\usepackage{colortbl}

\definecolor{ourcolor}{RGB}{235,245,255}
\definecolor{bestcolor}{RGB}{255,230,180}
\definecolor{sndcolor}{RGB}{225,240,255}
\definecolor{trdcolor}{RGB}{235,235,235}

\usepackage{cite}

\makeatletter
\let\NAT@parse\undefined
\makeatother

\usepackage{hyperref}

\title{\LARGE \bf
PiLoT v2: Pixel-to-Orthogonal Map Alignment for Free-view UAV Geo-localization
}

\author{
    Xinyi Liu$^{1}$ \quad Xiaoya Cheng$^{1}$ \quad Rouwan Wu$^{1}$ \quad Zhaochen Wang$^{1}$ \\
    Shen Yan$^{1}$ \quad Maojun Zhang$^{1}$ \quad Yu Liu$^{1\dagger}$ \\[4pt]
    {\small $^{1}$National University of Defense Technology} \\
    {\tt\small \{liuxinyi24, chengxy, wurouwan97, wangzhaochen, yanshen12, mjzhang, jasonyuliu\}@nudt.edu.cn} \\
}

\begin{document}

\maketitle

\thispagestyle{empty}
\pagestyle{empty}

\begin{abstract}
Real-time, drift-free UAV geo-localization is essential for autonomous missions in GNSS-denied environments. The pioneering system, PiLoT, achieves high precision via Neural Pixel-to-3D Registration, aligning UAV video streams with a single rendered reference view from 3D meshes. However, its reliance on heavy 3D meshes incurs massive storage overheads, complex map acquisition, and significant computational rendering costs, severely hindering deployment on embedded platforms. To address these bottlenecks, we propose PiLoT v2, a lightweight yet robust evolution that shifts the paradigm to direct pixel-to-orthogonal map registration for free-view UAV geo-localization. By leveraging True Digital Orthophoto Maps (TDOMs) and Digital Surface Models (DSMs) as the reference substrate, PiLoT v2 replaces GPU-intensive 3D rendering with a highly efficient, CPU-friendly map cropping operation. To bridge the severe geometric discrepancy between these 2.5D orthogonal crops and free-view oblique UAV imagery, we train a cross-view feature registration network using a novel, large-scale geometrically annotated dataset. Furthermore, we integrate onboard sensor priors—specifically gravity direction and single-point laser range—directly into the pose optimization manifold to enhance robustness against cross-view visual degradation. Experimental results demonstrate that PiLoT v2 achieves performance comparable to, or even exceeding, its Pixel-to-3D predecessor, while offering drastically lower storage and computational costs. 
\end{abstract}

\section{Introduction}

\begin{figure*}[t]
    \centering
    \includegraphics[width=\textwidth]{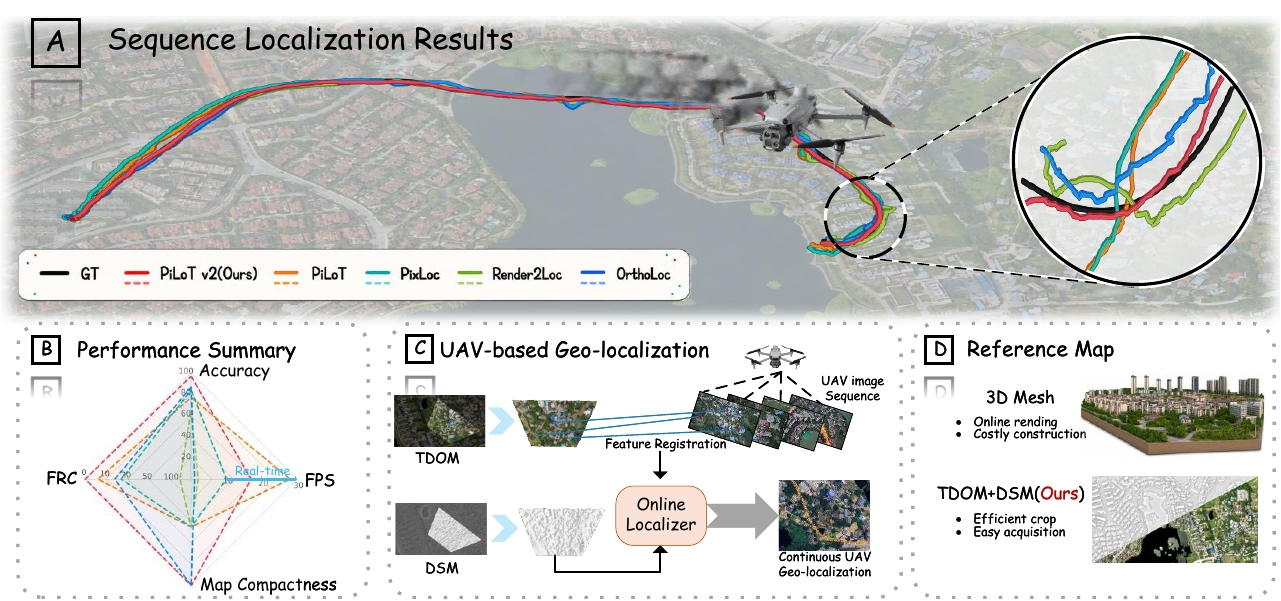}
    \caption{
    \textbf{Revisiting of sequence localization.}
    PiLoT v2 estimates globally aligned 6-DoF UAV poses from UAV sequences using lightweight DOM+DSM map crops.
    The radar chart shows that PiLoT v2 achieves a favorable balance for real-time onboard geo-localization in GNSS-denied environments, where FRC denotes failure recovery count.
    }
    \label{fig:teaser}

\end{figure*}

Unmanned Aerial Vehicle (UAV) geo-localization aims to estimate globally aligned 6-DoF poses for each frame in a UAV query sequence. 
Drift-free geo-localization is a fundamental prerequisite for pushing UAVs towards full autonomy in demanding real-world applications, such as digital twins, infrastructure inspection, augmented reality, and embodied AI\cite{qin2018vinsmono,campos2021orbslam3}.
This requirement becomes especially critical when GNSS signals are degraded or lost, where the last reliable GNSS/GPS estimate before signal loss can provide only a coarse initialization.

Introducing global map priors provides an effective way to anchor UAV trajectories to absolute geographic coordinates and eliminate accumulated drift. Recent high-accuracy UAV localization systems, such as Render2Loc~\cite{yan2023rendercompare} and PiLoT~\cite{cheng2026pilot}, formulate geo-localization as a Neural Pixel-to-3D Registration problem over geo-referenced 3D meshes. To maintain real-time performance, these methods typically align the video stream with a single rendered reference view. However, these 3D meshes demand massive storage (often $>10$~GB) and depend on costly oblique photogrammetry followed by time-intensive 3D reconstruction. Furthermore, online reference generation relies on computationally expensive rendering engines, placing a heavy burden on embedded GPU resources.

An alternative line of work investigates lightweight orthographic maps, such as True Digital Orthophoto Maps (TDOMs) and Digital Surface Models (DSMs), as replacements for heavy 3D mesh representations~\cite{kinnari2021gnssdenied,dhaouadi2026ortholoc}. These 2.5D maps are easier to acquire, straightforward to update, and significantly more compact. However, bridging the severe cross-view geometric discrepancies between pure orthographic maps and free-view oblique UAV images remains challenging. Most existing methods address this by relying on costly explicit image-to-map matching, making it difficult to support real-time sequential UAV geo-localization.

To break this performance bottleneck, we propose \textbf{PiLoT v2}. As visualized in Fig.~\ref{fig:teaser}, PiLoT v2 revisits the dual-thread localization paradigm of its predecessor and replaces the heavy 3D mesh representation with a lightweight TDOM+DSM map. Crucially, we shift the paradigm by replacing the exorbitant 3D rendering module with a highly efficient, CPU-friendly map cropping operation. As a result, PiLoT v2 preserves the real-time dual-thread structure for UAV geo-localization while being vastly more suitable for resource-constrained onboard deployment.

Using 2D orthographic crops inevitably introduces a massive cross-view gap when compared with oblique UAV query images~\cite{zheng2020university1652,yang2021l2ltr}. To overcome this, we train a cross-view feature registration network using a newly constructed, large-scale synthetic dataset with precise geometric annotations. This enables the network to extract highly consistent features despite drastic viewpoint differences, demonstrating strong zero-shot generalization to real-world UAV-to-orthophoto queries. Furthermore, built-in UAV sensor priors, specifically gravity direction and single-point laser range, are readily available during flight~\cite{sweeney2015gravity,ding2021gravity,yan2023longterm}. To mitigate visual degradation and improve robustness against unstable cross-view registration, we incorporate these physical constraints directly into the Levenberg-Marquardt (LM) based pose refinement manifold. Extensive experiments show that our method matches or even outperforms SoTA 3D-mesh-based methods, while using a drastically simplified TDOM+DSM map prior.

In summary, our main contributions are:
\begin{itemize}
    \item We propose PiLoT v2, a real-time UAV geo-localization pipeline that replaces costly 3D rendering with lightweight TDOM+DSM map cropping, dramatically reducing storage and computational overhead.
    \item We train a robust feature registration network using large-scale geometrically annotated cross-view data, bridging the modality gap between orthographic map crops and oblique UAV images.
    \item We integrate onboard gravity direction and single-point laser range priors directly into the LM-based pose optimization manifold to ensure robust cross-view registration.
    \item We construct a comprehensive TDOM+DSM benchmark aligned with corresponding 3D oblique meshes. Evaluations demonstrate that our lightweight method matches or surpasses SoTA 3D-mesh-based systems for real-time edge deployment.
\end{itemize}

\section{Related Work}
\subsection{UAV Geo-localization}
% \label{sec:references}
UAV geo-localization aims to estimate the 6-DoF pose of a UAV camera in a global coordinate frame. 
Since simultaneous localization and mapping (SLAM)~\cite{campos2021orbslam3} and visual-inertial odometry (VIO)~\cite{qin2018vinsmono} accumulate drift without global references, many methods register UAV images against geo-referenced maps. 
Early methods mainly rely on 2D satellite imagery~\cite{patel2020googleearth,sarlin2023orienternet}. 
These methods can provide coarse global localization, but they are usually limited to 2D or 3-DoF pose estimation under simplified orthographic assumptions. 
To recover full 6-DoF poses, recent works turn to simulation-driven references and geo-referenced 3D maps~\cite{yin2023isimloc,yan2023rendercompare,wu2024uavd4l,cheng2026pilot}. 
These methods typically initialize the pose using retrieval~\cite{hausler2021patchnetvlad,berton2024meshvpr}, sensor priors, or noisy pose estimates, and then refine it through feature matching~\cite{sun2021loftr,lindenberger2023lightglue,zhou2021patch2pix}, render-and-compare, or direct alignment~\cite{niu2025hgsloc,kerbl2023gaussian,huang2025sparse2dense}.
However, they usually require dense 3D reconstruction, large map storage, and online rendering to generate viewpoint-consistent references.
In contrast, our work achieves real-time sequential UAV geo-localization using lightweight TDOM/DSM crops instead of rendered 3D references.

\subsection{Lightweight Maps Representations}
% \label{sec:geodata}
To reduce the cost of 3D-mesh-based localization, recent studies explore lightweight map priors, including topographic maps~\cite{chen2021realtime}, orthographic imagery and orthophotos~\cite{kinnari2022season}, LoD city models~\cite{zhu2024lodloc}, and 2.5D aerial references such as TDOM/DSM geodata~\cite{dhaouadi2026ortholoc,ye2025anyvisloc}.
LoD-Loc~\cite{zhu2024lodloc} estimates UAV pose from coarse LoD maps, avoiding heavy textured reconstruction but still relying on projected 3D geometry and coarse sensor initialization. 
OrthoLoC~\cite{dhaouadi2026ortholoc} further shows that TDOM/DSM can support UAV 6-DoF geo-localization, while highlighting the projection discrepancy between perspective UAV images and orthographic references. 
AnyVisLoc\cite{ye2025anyvisloc} further highlights the difficulty of low-altitude multi-view UAV geo-localization under 2.5D aerial and satellite references. These works demonstrate the potential of lightweight maps. 
Our work follows this direction while targeting real-time geo-localization from continuous UAV image sequences using direct TDOM/DSM crops.

\subsection{Cross-view Registration}
% \label{sec:Registration}
Cross-view discrepancy is a key challenge in UAV geo-localization, especially when oblique UAV images are matched orthographic references~\cite{chen2021realtime,patel2020googleearth,kinnari2022season,dhaouadi2026ortholoc}. 
Existing works address this issue with cross-view retrieval~\cite{ahn2021contrastive,shi2022beyond}, feature matching~\cite{sun2021loftr,edstedt2023dkm,edstedt2024roma}, contrastive learning~\cite{ahn2021contrastive}, and geometry-aware supervision~\cite{wang2024dust3r,leroy2024mast3r}. 
Recent studies further show that synthetic data with accurate geometric annotations improves robustness to large viewpoint changes and supports sim-to-real transfer~\cite{cheng2026pilot,yin2023isimloc}. 
Compared with rendered references, our TDOM/DSM crops introduce a larger gap to oblique UAV queries. 
Therefore, we combine cross-view registration with built-in UAV sensor priors, including gravity direction and single-point laser range, for more robust pose refinement.

\section{Dataset and Benchmark}
\label{sec:dataset_Benchmark}
\subsection{Cross-view Training Data Generation}
\label{sec:training_generation}
This section describes the generation and processing pipeline of the training data.
The data are used to train the feature registration network between oblique UAV images and orthographic reference images.

\begin{figure}[t]
    \vspace*{8pt}
    \centering
    \includegraphics[width=\columnwidth]{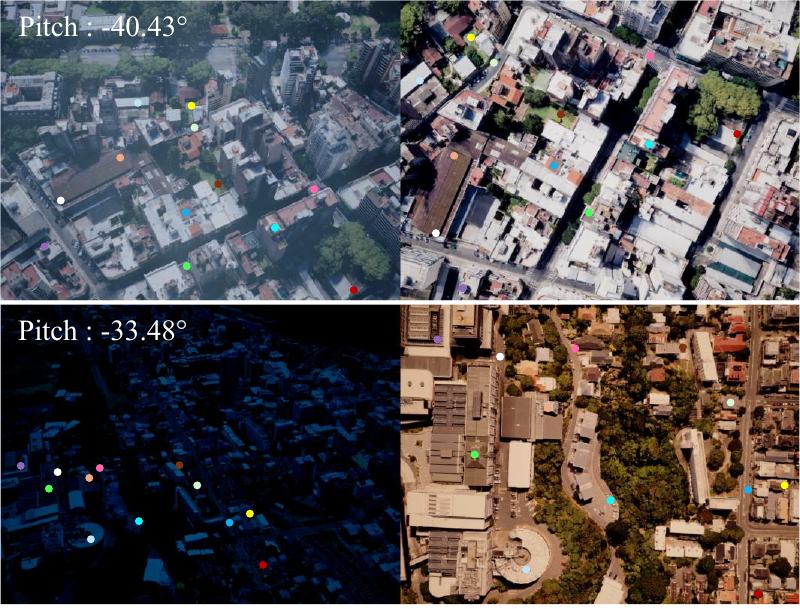}
    \caption{
    \textbf{Visualization of complementary training pairs.}
    The top row shows an oblique-to-oblique pair from the original PiLoT training set, while the bottom row shows an oblique-to-orthographic pair from AirZoo.
    Yellow dots indicate geometrically verified matched points used for feature-metric training.
    }
    \label{fig:supp_training_data}
\end{figure}

\noindent\textbf{Data Generation.}
The data generation pipeline is built on the \textbf{AirSim--Cesium--Unreal Engine platform}. 
For each scene, we load geo-referenced 3D maps and simulate UAV flights along predefined WGS84 waypoints, from which RGB images, pixel-wise depth maps, camera intrinsics, 6-DoF poses, and geo-coordinate annotations are recorded. 
We render both oblique and orthographic sequences within the same scene coordinate system, oblique views serve as UAV query images, while orthographic views provide map-side references. 
By varying regions, weather, illumination, flight altitude, and camera pitch, the generated data cover diverse appearances and UAV viewpoints, as shown in Fig.~\ref{fig:supp_training_data}

\noindent\textbf{Data Process.}
Based on the depth maps, camera intrinsics, and 6-DoF poses, we compute the visible-area overlap between oblique and orthographic images to construct candidate query-reference pairs. 
For each oblique query image, orthographic references are ranked by overlap score, while a maximum usage count is imposed on each reference image to avoid sample imbalance. 
We further filter the candidates with geometric consistency checks, removing pairs with insufficient overlap or unstable projection. 
The resulting reliable oblique-to-orthographic pairs are mixed with the original PiLoT training data for network training.
% The statistics of the two training data sources are summarized in Tab.~\ref{tab:training_data}.

\subsection{Evaluation Benchmark}
\label{sec:test_benchmark}
We evaluate UAV geo-localization performance on three benchmark groups following the test protocol of PiLoT~\cite{cheng2026pilot}.
The original benchmark provides UAV query sequences, ground-truth poses, and geo-referenced 3D meshes, but it does not provide TDOM/DSM maps.
We construct a comprehensive TDOM+DSM dataset aligned with the corresponding 3D meshes for all test scenes.

\begin{figure}[t]
    \vspace*{8pt}
    \centering
    \includegraphics[width=\columnwidth]{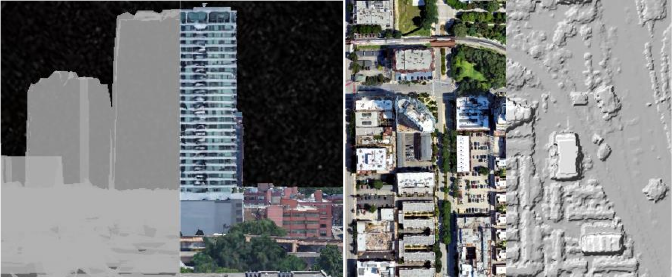}
    \caption{
    \textbf{Visualization of the geographic alignment.}
    The left panel shows the alignment between the generated DSM and the original map, with the DSM displayed on the left and the original map on the right.
    The right panel shows the geographic consistency between the generated DOM and DSM, with the DOM on the left and the DSM on the right.
    }
    \label{fig:map_generation}
\end{figure}

\begin{table}[t]
\vspace*{8pt}
\centering
\scriptsize
\caption{Statistics of the constructed TDOM/DSM evaluation maps.}
\label{tab:dataset_stats}
\setlength{\tabcolsep}{4pt}
\renewcommand{\arraystretch}{1.05}
\setlength{\cmidrulewidth}{0.25pt}
\resizebox{\columnwidth}{!}{
\begin{tabular}{@{}c@{\hspace{2pt}}lccccc@{}}
\toprule
\multicolumn{2}{c}{Scene} & Img. & Area & Type & Category & City \\
\midrule

\multirow{5}{*}{SynthCity-6}
& Switzerland-4  & 4.5k  & 1.44 km$^2$ & Synthetic & Rural & Thun \\
\cmidrule(r){2-7}
& Switzerland-7  & 4.5k  & 1.84 km$^2$ & Synthetic & Rural & Bern \\
\cmidrule(r){2-7}
& Switzerland-12 & 4.5k  & 1.43 km$^2$ & Synthetic & Rural & Lausanne \\
\cmidrule(r){2-7}
& USA-2          & 4.5k  & 2.25 km$^2$ & Synthetic & Urban & Chicago \\
\cmidrule(r){2-7}
& USA-8          & 4.5k  & 2.54 km$^2$ & Synthetic & Urban & Cupertino \\
\midrule

\multirow{2}{*}{UAVScenes}
& AMvalley  & 11.3k & 0.84 km$^2$ & Real & Valley & HonKong \\
\cmidrule(r){2-7}
& HKairport & 7.2k  & 0.28 km$^2$ & Real & Airport & HongKong\\
\midrule

\multicolumn{2}{c}{UAVD4L-2yr}
& 6.3k & 4.17 km$^2$ & Real & Rural \& Urban & Changsha \\
\bottomrule
\end{tabular}
}
\end{table}

\noindent\textbf{Benchmark.}
As summarized in Tab.~\ref{tab:dataset_stats}, the constructed TDOM/DSM dataset is built from selected test scenes in three benchmark groups.
We use SynthCity-6~\cite{cheng2026pilot} to study localization accuracy in controlled synthetic environments. SynthCity-6 is a synthetic UAV localization benchmark, providing consistent scene geometry.
UAVScenes~\cite{wang2025uavscenes} is a real-world UAV localization benchmark collected in outdoor environments, introducing realistic imaging conditions, scene geometry. 
UAVD4L-2yr contain seasonal, illumination, and scene-content changes, where query flights and reference maps exhibit large temporal gaps. 

\begin{figure*}[t]
    \vspace*{8pt}
    \centering
    \includegraphics[width=\textwidth]{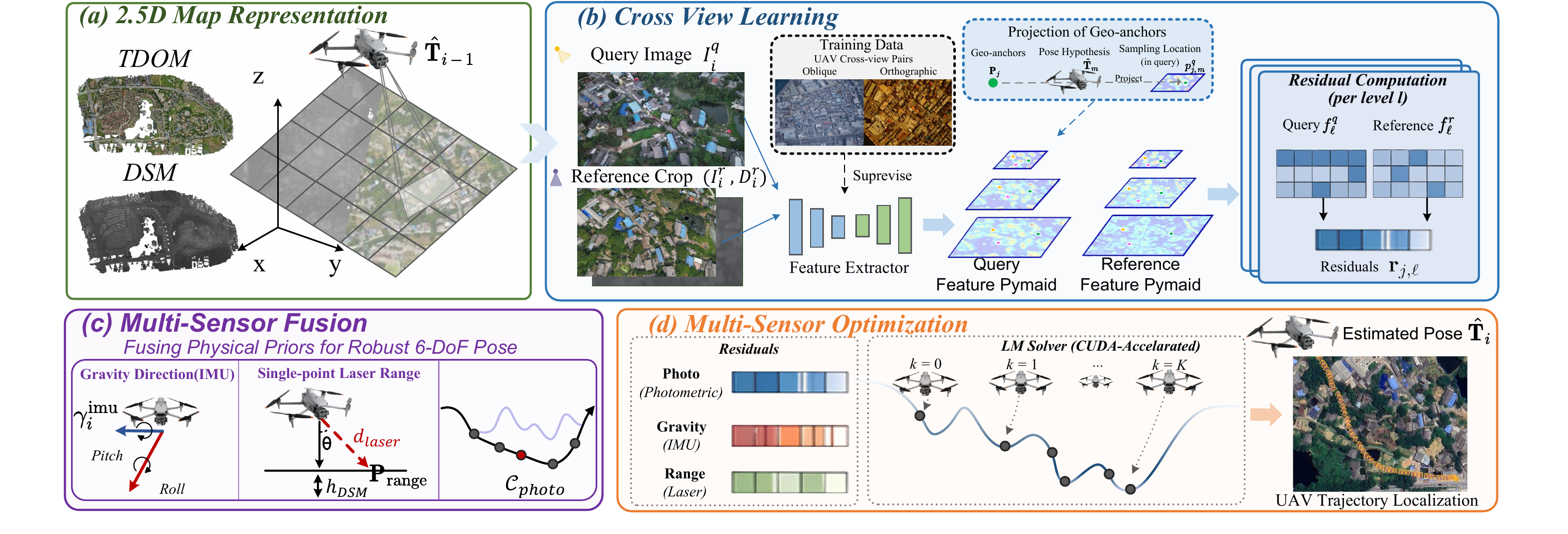}
    \caption{
    \textbf{Overview of the proposed framework.}
    (a) A local reference view $(I_i^r,D_i^r)$ is cropped from the lightweight TDOM/DSM maps according to the pose prior.
    (b) Reference is jointly encoded with UAV query image to compute cross-view photo residuals.
    (c) Gravity direction and single-point laser range measurements are fused as built-in-UAV sensor residuals.
    (d) The photo and sensor residuals are jointly minimized by a coarse-to-fine LM optimizer, producing a refined globally aligned 6-DoF UAV pose.
    }
\label{fig:method}
\end{figure*}

\noindent\textbf{2.5D Map Construction.}
For each scene, we generate geographically aligned TDOM/DSM maps as the reference representation for crop-based localization. Fig.~\ref{fig:map_generation} visualizes the spatial consistency.
Despite the original maps are stored in different formats, including mesh-format scenes for SynthCity-6 and UAVD4L-2yr and PLY-format scenes for UAVScenes, they are processed with the same map-generation procedure. 
Specifically, we define an orthographic trajectory over each scene, render overlapping images with known camera poses, and reconstruct a dense surface from these observations. 
The visible surface height is rasterized into a DSM, while the texture observations are orthorectified onto the same grid to obtain the TDOM. 
The constructed maps cover 14.78 km$^2$ in total and are generated offline. This cost is not included in the online localization runtime.

\section{Method}
\label{sec:method}
Given geo-referenced TDOM/DSM maps, a monocular UAV video stream $\{I_i^q\}$ with known intrinsics $\{\mathbf{K}_i\}$, and a initial prior$\tilde{\mathbf{T}}_{\mathrm{init}}$, our objective is to sequentially estimate the absolute 6-DoF pose $\hat{\mathbf{T}}_i$ for each oblique query image $I_i^q$ in the global coordinate frame of the map.
The initial prior can be provided by the last reliable GNSS estimate before signal degradation or loss.
Fig.~\ref{fig:method} shows the overall pipeline.

\subsection{Preliminary}
We follow the pixel-to-3D registration pipeline used in PiLoT~\cite{cheng2026pilot}.
PiLoT renders an oblique reference image from a geo-referenced 3D mesh according to a pose prior. Its key idea is to convert the reference into pixel-aligned 3D geo-anchors, and then directly register query pixels to these 3D anchors. The pose is refined by minimizing the learned photo residuals between the projected query pixels and the corresponding reference pixels.

\subsection{2.5D Map Representation}
\label{sec:representation}
Compared with geo-referenced 3D mesh used in render-based localization, TDOM/DSM maps are easier to obtain from satellite products or standard orthographic aerial surveys.
They preserve the two types of information required by our geo-localization backend. TDOM provides geo-referenced appearance cues, while DSM provides metric height values for constructing 3D geo-anchors.
As illustrated in Fig.~\ref{fig:method}(a), for the current query image $I_i^q$, we use the pose estimated from the previous query image $\hat{\mathbf{T}}_{i-1}$ as the pose prior to crop a local reference view $(I_i^r,D_i^r)$ from the TDOM/DSM maps.
The resulting reference view contains the TDOM reference image and the pixel-aligned DSM height map.

\subsection{Cross-View Learning}
\label{sec:Learning}
Our geo-localization pipeline operates by registering the query image $I_i^q$ against a local 2.5D reference view $(I_i^r, D_i^r)$, which is shown in Fig.~\ref{fig:method}(b).
This process is formulated as an end-to-end neural-guided optimization that bridges photo consensus and physical constraints.

To process the drastic viewpoint difference between the oblique query and the orthographic reference, we utilize the proven feature extraction backbone from PiLoT.
A lightweight encoder-decoder architecture extracts multi-scale feature pyramids and uncertainty maps $\{(\mathbf{f}_{\ell}^{q}, w_{\ell}^{q}),(\mathbf{f}_{\ell}^{r}, w_{\ell}^{r})\}_{\ell=0}^{2}$ for the query image and the reference image, where $\ell$ indexes the feature pyramid level.
For a sampled reference pixel $\mathbf{p}_{j}^{r}$, its DSM value $D_i^r(\mathbf{p}_{j}^{r})$ lifts it to a 3D geo-anchor $\mathbf{P}_{j}$.
At pyramid level $\ell$, the photo residual of the $m$-th pose hypothesis $\tilde{\mathbf{T}}_m$ is defined as
\begin{equation}
\mathbf{r}_{j,\ell}^{(m)}
=
\mathbf{f}_{\ell}^{q}
\left(
\pi
\left(
\mathbf{K}_{\ell},
\tilde{\mathbf{T}}_{m}^{-1},
\mathbf{P}_{j}
\right)
\right)
-
\mathbf{f}_{\ell}^{r}
\left(
\mathbf{p}_{j}^{r}
\right),
\label{eq:photo_residual}
\end{equation}
where $\pi(\cdot)$ denotes the pinhole projection function and $\mathbf{K}_{\ell}$ is the intrinsic matrix scaled to the $\ell$-th feature level.
The photo cost aggregates these residuals using a Huber robust loss $\rho(\cdot)$~\cite{huber1992robust}:
\begin{equation}
\mathcal{C}_{\mathrm{photo}}^{(m,\ell)}
=
\sum_{j}
\rho
\left(
w_{\ell}(j)
\left\|
\mathbf{r}_{j,\ell}^{(m)}
\right\|_2^2
\right),
\label{eq:photo_cost}
\end{equation}

\subsection{Multi-Sensor Prior Fusion}
\label{sec:fusion}
The lightweight 2.5D reference introduces ambiguity because TDOM/DSM maps lack side-view texture and oblique appearance cues.
To regularize pose estimation, we incorporate two built-in UAV sensor residuals: an IMU gravity direction residual and a single-point laser range residual.

\noindent\textbf{IMU Gravity Direction Residual.}
The IMU gravity direction residual is obtained from Inertial Measurement Unit (IMU), which provides a strong constraint on the UAV attitude.
The built-in-UAV IMU provides a gravity direction measurement $\boldsymbol{\gamma}_{i}^{\mathrm{imu}}$ for the current query image.
Given the $m$-th pose hypothesis, its rotation $\mathbf{R}(\tilde{\mathbf{T}}_m)$ predicts the gravity direction in the camera frame, yielding the IMU residual
\begin{equation}
\mathbf{r}_{\mathrm{imu}}^{(m)}
=
\mathbf{R}(\tilde{\mathbf{T}}_m)^{\top}
\boldsymbol{\gamma}^{W}
-
\boldsymbol{\gamma}_{i}^{\mathrm{imu}},
\label{eq:imu_residual}
\end{equation}
where $\boldsymbol{\gamma}^{W}$ is the gravity direction in the world coordinate frame.

\noindent\textbf{Single-point Laser Range Residual.}
The single-point laser range residual provides a direct distance measurement from the UAV to the observed surface.
The laser range measurement $d_i^{\mathrm{range}}$ is associated with the center pixel $\mathbf{p}_{i,c}^{q}=[u_{i,c},v_{i,c},1]^{\top}$ of the current query image.
The corresponding unit ray in the camera frame is
\begin{equation}
\mathbf{v}_{i,c}
=
\frac{
\mathbf{K}_i^{-1}\mathbf{p}_{i,c}^{q}
}{
\left\|
\mathbf{K}_i^{-1}\mathbf{p}_{i,c}^{q}
\right\|_2
}.
\label{eq:camera_ray}
\end{equation}
Using the rotation $\mathbf{R}(\tilde{\mathbf{T}}_m)$ and camera center $\mathbf{C}(\tilde{\mathbf{T}}_m)$ of the pose hypothesis, the range-induced 3D point is
\begin{equation}
\mathbf{P}_{\mathrm{range}}^{(m)}
=
\mathbf{C}(\tilde{\mathbf{T}}_m)
+
\mathbf{R}(\tilde{\mathbf{T}}_m)
\left(
d_i^{\mathrm{range}}\mathbf{v}_{i,c}
\right).
\label{eq:range_point}
\end{equation}
Letting $\mathbf{P}_{\mathrm{range}}^{(m)}=(x_{\mathrm{range}}^{(m)},y_{\mathrm{range}}^{(m)},z_{\mathrm{range}}^{(m)})$, the range residual is
\begin{equation}
r_{\mathrm{range}}^{(m)}
=
z_{\mathrm{range}}^{(m)}
-
D_i^r
\left(
x_{\mathrm{range}}^{(m)},
y_{\mathrm{range}}^{(m)}
\right).
\label{eq:range_residual}
\end{equation}
This residual anchors the pose to the local DSM surface and reduces depth ambiguity in the 2.5D reference.

\subsection{Multi-Sensor Optimization}
\label{sec:optimir}
\begin{figure}[t]
    \vspace*{8pt}
    \centering
    \includegraphics[width=\columnwidth]{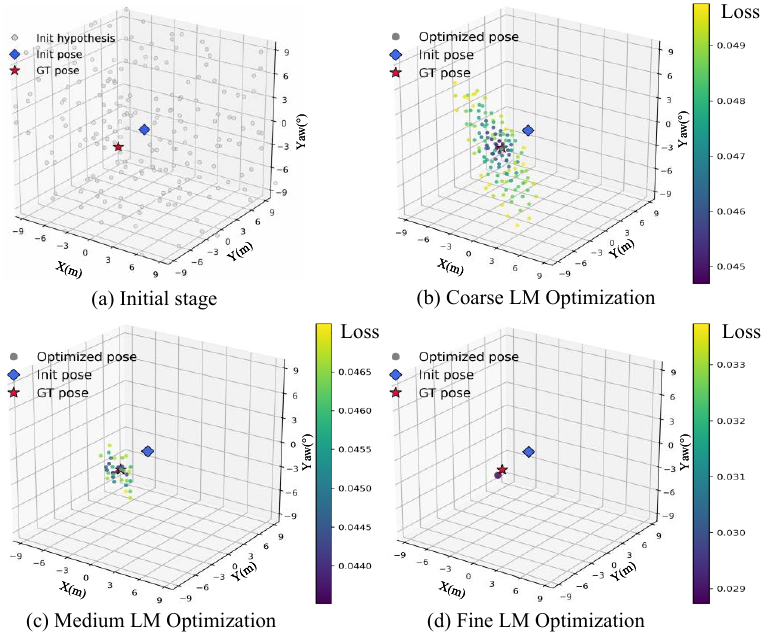}
    \caption{
    \textbf{Multi-hypothesis coarse-to-fine LM optimization.}
    We sample initial pose hypotheses in the local $(x,y,\psi)$ space and refine them through coarse, medium, and fine LM optimization.
    }
    \label{fig:lm_optimization}
\end{figure}

Given the photo residuals from cross-view feature registration and the sensor residuals defined in Sec.~\ref{sec:fusion}, we refine the query pose with a multi-hypothesis Levenberg--Marquardt (LM) optimizer.
Since TDOM/DSM reference views are generated from a pose prior, errors in horizontal translation and yaw may lead to an incorrect map crop and degrade the subsequent registration.
To improve robustness against such initialization errors, we sample multiple pose hypotheses $\{\tilde{\mathbf{T}}_m\}_{m=1}^{M}$ in the local $(x,y,\psi)$ space around the propagated pose prior $\hat{\mathbf{T}}_{i|i-1}$, where $x$ and $y$ denote horizontal translation offsets and $\psi$ denotes yaw.
As illustrated in Fig.~\ref{fig:lm_optimization}(a), these initial hypotheses cover a local uncertainty region around the input pose prior, providing a wider convergence basin before LM refinement.

For each pose hypothesis, the final refinement objective is formulated as
\begin{equation}
\begin{aligned}
\mathcal{C}_{\mathrm{total}}^{(m)}
=
&
\mathcal{C}_{\mathrm{photo}}^{(m,\ell)}
+
\lambda_{\mathrm{imu}}
\rho
\left(
\left\|
\mathbf{r}_{\mathrm{imu}}^{(m)}
\right\|_2^2
\right)
\\
&
+
\lambda_{\mathrm{range}}
\rho
\left(
\left(
r_{\mathrm{range}}^{(m)}
\right)^2
\right),
\end{aligned}
\label{eq:total_objective}
\end{equation}
where $\lambda_{\mathrm{imu}}$ and $\lambda_{\mathrm{range}}$ balance the IMU gravity direction residual and the single-point laser range residual.
The photo residual drives the cross-view registration between the query image and the TDOM/DSM reference view, while the IMU and range residuals regularize the attitude and vertical geometry.
The loss values visualized in Fig.~\ref{fig:lm_optimization} correspond to this joint objective, where lower values indicate hypotheses that better satisfy both photo registration and sensor constraints.

We perform the optimization in a coarse-to-fine manner over three feature pyramid levels, as shown in Fig.~\ref{fig:lm_optimization}(b)--(d).
At each LM iteration, all residuals are linearized around the current pose hypothesis and assembled into a damped normal equation:
\begin{equation}
\left(
\mathbf{J}^{\top}
\mathbf{W}
\mathbf{J}
+
\lambda
\mathbf{I}
\right)
\Delta\boldsymbol{\xi}
=
-
\mathbf{J}^{\top}
\mathbf{W}
\mathbf{r},
\label{eq:lm_normal_equation}
\end{equation}
where $\mathbf{J}$ is the Jacobian matrix, $\mathbf{W}$ is the diagonal uncertainty weight matrix, $\lambda$ is the LM damping factor, and $\Delta\boldsymbol{\xi}$ is the 6-DoF pose increment.
The pose hypothesis is updated on $SE(3)$ by
\begin{equation}
\tilde{\mathbf{T}}_{m}^{(k+1)}
=
\exp
\left(
\Delta\boldsymbol{\xi}
\right)
\cdot
\tilde{\mathbf{T}}_{m}^{(k)} .
\label{eq:pose_update}
\end{equation}

The coarse level first evaluates the sampled hypotheses over a wider basin and suppresses clearly inconsistent hypotheses.
The medium level then refines the remaining hypotheses with more discriminative feature responses.
Finally, the fine level performs high-resolution pose refinement, where the best hypothesis converges close to the ground-truth pose with the lowest objective value.
After optimization, the hypothesis with the lowest final objective value is selected as the output pose $\hat{\mathbf{T}}_i$ of the current query image.

\subsection{Cross-View Geometry Training Strategy}
\label{sec:training}
Although PiLoT learns geometry-aware UAV features from large-scale synthetic training data, its pretrained weights are mainly learned from oblique-to-oblique training data.
Directly applying them to the registration between oblique UAV images and orthographic reference images leads to a viewpoint-distribution gap.
The generation of our cross-view training data is detailed in Sec.~\ref{sec:training_generation}.

% =========================
% Comprehensive Localization Results
% =========================
\begin{table*}[t]
\vspace*{8pt}
\centering
\scriptsize
\setlength{\tabcolsep}{3pt}
\renewcommand{\arraystretch}{1.05}
\caption{
Comprehensive UAV geo-localization results.
We compare our method with 3D render-based methods and TDOM/DSM crop-based baselines on synthetic and real-world datasets. Median errors are reported in meters (m) and degrees ($^\circ$).
}
\label{tab:comprehensive_results}
\resizebox{\textwidth}{!}{
\begin{tabular}{l c ccc ccc ccc}
\toprule
\multirow{2}{*}{Method}
& \multirow{2}{*}{FPS$\uparrow$}
& \multicolumn{3}{c}{SynthCity-6}
& \multicolumn{3}{c}{UAVScenes}
& \multicolumn{3}{c}{UAVD4L-2yr} \\
\cmidrule(lr){3-5}
\cmidrule(lr){6-8}
\cmidrule(lr){9-11}
&
& Med.m$\downarrow$ / Med.$^\circ\downarrow$
& R@1/3/5/10$\uparrow$
& FRC$\downarrow$
& Med.m$\downarrow$ / Med.$^\circ\downarrow$
& R@1/3/5/10$\uparrow$
& FRC$\downarrow$
& Med.m$\downarrow$ / Med.$^\circ\downarrow$
& R@1/3/5/10$\uparrow$
& FRC$\downarrow$ \\
\midrule

\multicolumn{11}{l}{\textit{3D Render-based}} \\

Render2Loc-eLoFTR
& 2
& 1.150 / 0.175
& 48.55 / 87.19 / 96.06 / 99.62
& 21
& 3.763 / 1.503
& 1.86 / 50.98 / 76.85 / 85.00
& 2399
& 2.251 / 1.863
& 6.73 / 61.82 / 76.89 / 87.14
& 232 \\

Render2Loc-RoMa
& 1
& \textbf{0.592} / \textbf{0.088}
& 73.72 / 96.36 / 99.28 / 99.97
& 5
& 3.641 / 1.413
& 2.51 / 53.91 / 76.35 / 91.05
& 1159
& 4.222 / 2.022
& 8.98 / 51.81 / 65.19 / 78.30
& 147 \\

PixLoc
& 10
& 0.759 / 0.374
& 69.81 / 93.81 / 94.71 / 99.17
& 5
& 3.414 / 1.176
& 4.14 / 53.00 / 81.44 / 98.78
& 5
& 2.201 / 1.794
& 7.75 / 67.57 / 83.82 / 94.49
& 17 \\

PiLoT
& \textbf{28}
& 0.731 / 0.109
& \textbf{74.51} / 98.66 / 98.87 / 98.94
& \textbf{0}
& 2.955 / 1.124
& \textbf{7.22} / \textbf{67.78} / 81.85 / 98.65
& 5
& 2.251 / 1.863
& 6.73 / 61.82 / 76.89 / 87.14
& 6 \\

\midrule

\multicolumn{11}{l}{\textit{TDOM/DSM Crop-based}} \\

OrthoLoC-eLoFTR
& 8
& 0.924 / 0.139
& 57.93 / 96.78 / 99.88 / \textbf{100.00}
& \textbf{0}
& 4.317 / 1.358
& 0.90 / 24.07 / 65.40 / 96.92
& 64
& 3.146 / 1.940
& 5.13 / 51.45 / 74.84 / 88.54
& 342 \\

OrthoLoC-RoMa
& 1
& 0.914 / 0.144
& 58.48 / 96.00 / 99.60 / \textbf{100.00}
& \textbf{0}
& 4.440 / 1.371
& 1.00 / 22.74 / 64.93 / 96.85
& 162
& 2.813 / 1.621
& 9.75 / 70.64 / 86.75 / 97.26
& 14 \\

\midrule

\rowcolor{ourcolor}
\textbf{Ours}
& 17
& 0.778 / 0.124
& 71.31 / \textbf{99.69} / \textbf{99.99} / \textbf{100.00}
& \textbf{0}
& \textbf{2.849} / \textbf{0.415}
& 3.90 / 58.25 / \textbf{99.75} / \textbf{99.91}
& \textbf{0}
& \textbf{1.577} / \textbf{1.488}
& \textbf{12.89} / \textbf{85.05} / \textbf{96.79} / \textbf{98.59}
& \textbf{0} \\

\bottomrule
\end{tabular}
}
\end{table*}
\begin{figure*}[t]
    \centering
    \includegraphics[width=\textwidth]{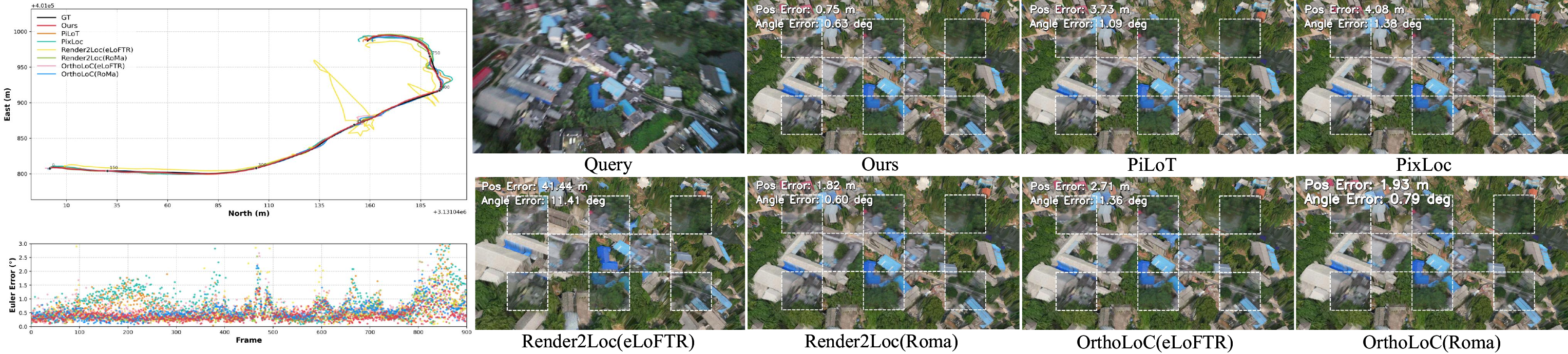}
    \caption{
    \textbf{Frame-wise localization error on a representative continuous UAV sequence.}
    We visualize the errors of different methods over a continuous UAV sequence.
    Compared with competing methods, our crop-based localization framework maintains a more stable trajectory and avoids large error spikes, indicating better long-term robustness under lightweight TDOM/DSM map priors.
    }
    \label{fig:trajectory_error}
\end{figure*}

\noindent\textbf{Supervision.}
We train the network using a direct registration approach.
For each training sample, the training objective minimizes a robust geometric loss $\mathcal{L}$ based on the reprojection error between the ground-truth query projection $\mathbf{p}_{j}^{q}$ and the estimated projection $\tilde{\mathbf{p}}_{j}^{q}$ of the same geo-anchor:
\begin{equation}
\mathcal{L}
=
\sum_{j}
\rho_B
\left(
\left\|
\mathbf{p}_{j}^{q}
-
\tilde{\mathbf{p}}_{j}^{q}
\right\|_2^2
\right),
\label{eq:training_loss}
\end{equation}
where $\rho_B(\cdot)$ is Barron's robust loss~\cite{barron2019robustloss}.
The uncertainty maps are also learned through this end-to-end supervision.
Specifically, they form the weighting matrix $\mathbf{W}$ in the differentiable optimizer, modulating the Jacobian $\mathbf{J}$ to solve the pose increment $\Delta\boldsymbol{\xi}$.
The updated pose then determines the sampling locations $\tilde{\mathbf{p}}_j^q$ used in ~\eqref{eq:training_loss}.
Therefore, although no explicit ground-truth uncertainty label is required, $w_\ell$ is implicitly supervised by the final geometric loss to down-weight unreliable anchors and suppress outliers during optimization.

\section{Experiments}
\label{sec:experiments}
\subsection{Experimental Setup}
\label{sec:experimental_setup}
\noindent\textbf{Implementation Details.}
We evaluate our crop-based localization framework on continuous UAV video sequences.
Both query images and reference crops are resized to 512 before feature extraction.
For pose refinement, we sample 500 valid 3D geo-anchors from the reference crop and perform a three-level coarse-to-fine LM optimization with 2, 3, and 4 iterations from coarse to fine.

For training, we adopt the same lightweight MobileOne-based~\cite{vasu2023mobileone} feature extraction architecture, and train it from scratch for orthographic-to-oblique feature registration.
The datasets consists of reference-query pairs sampled from generated UAV trajectories, where the query poses are perturbed with random noise of 5--15~m in translation and 5--15$^\circ$ in rotation to simulate pose-prior uncertainty.
The network is supervised by a geometric reprojection loss over 500 reference geo-anchors.
We train the model for 30 epochs using the Adam optimizer with an initial learning rate of $1\times10^{-3}$ on 8 RTX 4090 GPUs.
Runtime is measured on an RTX 5090 GPU.

\noindent\textbf{Datasets.}
We evaluate all methods on the constructed TDOM/DSM benchmark described in Sec.~\ref{sec:test_benchmark}.

\noindent\textbf{Baselines.}
We compare against two categories of baselines. The first category contains render-based localization methods that rely on 3D mesh references, including PiLoT~\cite{cheng2026pilot}, PixLoc~\cite{sarlin2021pixloc}, and Render2Loc~\cite{yan2023rendercompare}. For Render2Loc, we further evaluate two matching backends, including RoMa~\cite{yan2023rendercompare} and eLoFTR~\cite{wang2024eloftr}.
The second category contains orthographic maps localization baselines following the OrthoLoC~\cite{dhaouadi2026ortholoc} protocol. They localize the query image against the local TDOM/DSM crop generated by our crop pipeline. We evaluate OrthoLoC with the same matchers used in Render2Loc.

\noindent\textbf{Metrics.}
We report standard localization metrics for UAV geo-localization.
Translation and rotation accuracy are measured by median errors.
We also report Recall@1/3/5/10, which denotes the percentage of frames localized within 1~m/1$^\circ$, 3~m/3$^\circ$, 5~m/5$^\circ$, and 10~m/10$^\circ$, respectively.
FPS is reported to measure the online running efficiency of each method.
We additionally report Failure Recovery Count (FRC) to evaluate sequence-level stability.
When the pose error exceeds 20~m or 20$^\circ$, the next frame is reinitialized with the ground-truth pose and one recovery is counted.

\subsection{UAV-based Geo-localization}
\label{sec:ego_localization}

\noindent\textbf{Results.}
Tab.~\ref{tab:comprehensive_results} summarizes the localization results across all three datasets, and Fig.~\ref{fig:trajectory_error} visualizes frame-wise errors on a continuous oblique UAV sequence.
On SynthCity-6 benchmark, our method is competitive in median accuracy and achieves the best Recall@3/5/10 with zero FRC, indicating stable localization from TDOM/DSM crops.
On the more challenging UAVScenes and UAVD4L-2yr benchmarks, Our method achieves the best overall accuracy under real-world map-query discrepancies and long-term appearance changes.
With cross-view registration and built-in UAV sensor priors, the system still runs at 17 FPS without scene-specific fine-tuning, demonstrating a practical balance among accuracy, robustness, efficiency, map lightweightness, and zero-shot generalization.
With built-in UAV sensor priors, our method still runs at 17 FPS and achieves real-time performance, demonstrating a practical balance among accuracy, robustness, efficiency, map lightweightness, and zero-shot generalization.

\subsection{Ablation Study}
\label{sec:ablation}
We conduct four ablation studies to validate the key components of the proposed framework.

% =========================
% Ablation: Sensor Priors
% =========================
\begin{table}[t]
\vspace*{8pt}
\centering
\scriptsize
\caption{
Ablation study of built-in-UAV sensor priors. Our method combines both priors.
Range and Gravity denote the single-point laser range prior and gravity direction prior, respectively.
}
\label{tab:ablation_priors}
\resizebox{\columnwidth}{!}{
\begin{tabular}{l cc cc c c}
\hline
Setting
& Range
& Gravity
& Med.m$\downarrow$
& Med.$^\circ\downarrow$
& \textbf{R@1/3/5/10}~(m,$^\circ$)$\uparrow$
& FRC$\downarrow$ \\
\hline

Range only
& \checkmark
& --
& 3.052
& 2.016
& 3.40 / 49.02 / 73.17 / 90.98
& \textbf{0} \\

Gravity only
& --
& \checkmark
& 2.189
& 1.511
& 10.89 / 68.40 / 86.38 / 95.63
& 12 \\

\rowcolor{ourcolor}
\textbf{Ours}
& \checkmark
& \checkmark
& \textbf{1.577}
& \textbf{1.488}
& \textbf{12.89} / \textbf{85.05} / \textbf{96.79} / \textbf{98.59}
& \textbf{0} \\

\hline
\end{tabular}
}
\end{table}

% =========================
% Ablation: Cross-view Training
% =========================
\begin{table}[t]
\vspace*{8pt}
\centering
\scriptsize
\caption{
Ablation study of cross-view feature training. 
W/o denote without.
}
\label{tab:ablation_crossview}
\resizebox{\columnwidth}{!}{
\begin{tabular}{l cc cc c}
\hline
Setting
& Med.m$\downarrow$
& Med.$^\circ\downarrow$
& Std.m$\downarrow$
& Std.$^\circ\downarrow$
& \textbf{R@1/3/5/10}~(m,$^\circ$)$\uparrow$ \\
\hline

w/o cross-view training
& 2.438
& 0.994
& 1.478
& 0.566
& 15.64 / 68.75 / 86.20 / 98.58 \\

\rowcolor{ourcolor}
\textbf{Ours}
& \textbf{1.458}
& \textbf{0.941}
& \textbf{0.681}
& \textbf{0.488}
& \textbf{27.99} / \textbf{89.08} / \textbf{99.25} / \textbf{100.00} \\

\hline
\end{tabular}
}
\end{table}

% =========================
% Ablation: Map Acquisition Cost
% =========================
\begin{table}[t]
\vspace*{8pt}
\centering
\scriptsize
\caption{
Ablation study of reference-map acquisition cost on UAVD4L-2yr.
}
\label{tab:ablation_map_cost}
\resizebox{\columnwidth}{!}{
\begin{tabular}{l c c c c}
\hline
Map representation
& Reference acquisition
& Map size$\downarrow$
& Peak GPU memory$\downarrow$
& Time$\downarrow$ \\
\hline

3D mesh
& Rendering
& 19.5~GB
& 619~MiB
& \textbf{15~ms} \\

\rowcolor{ourcolor}
\textbf{TDOM/DSM}
& Crop
& \textbf{472~MB}
& \textbf{0~MiB}
& \textbf{13~ms} \\

\hline
\end{tabular}
}
\end{table}

% =========================
% Ablation: Map Prior Comparison
% =========================
\begin{table}[t]
\vspace*{8pt}
\centering
\scriptsize
\caption{
Comparison with geo-referenced 3D mesh on UAVD4L-2yr.
}
\label{tab:ablation_high_quality_map}
\resizebox{\columnwidth}{!}{
\begin{tabular}{l c c cc c c}
\hline
Method
& Map prior
& FPS$\uparrow$
& Med.m$\downarrow$
& Med.$^\circ\downarrow$
& \textbf{R@1/3/5/10}~(m,$^\circ$)$\uparrow$ \\
\hline

PiLoT
& geo-referenced 3D mesh
& \textbf{28}
& 1.600
& 1.582
& \textbf{15.82} / 84.47 / 93.24 / 97.89 \\

\rowcolor{ourcolor}
\textbf{Ours}
& TDOM/DSM crops
& 17
& \textbf{1.577}
& \textbf{1.488}
& 12.89 / \textbf{85.05} / \textbf{96.79} / \textbf{98.59} \\

\hline
\end{tabular}
}
\end{table}

\noindent\textbf{Effect of Sensor Priors.}
Tab.~\ref{tab:ablation_priors} shows that both built-in UAV sensor priors contribute to stable pose refinement.
The gravity direction prior mainly regularizes UAV attitude and helps suppress trajectory failures, while the single-point laser range prior constrains depth ambiguity with respect to the DSM surface and improves localization accuracy. 
Combining both priors achieves the best overall performance.

\noindent\textbf{Effect of Cross-view Training.}
Tab.~\ref{tab:ablation_crossview} shows that cross-view training is critical for crop-based localization.
Without it, the model suffers from a clear performance drop due to the orthographic-to-oblique viewpoint gap.
After introducing the proposed cross-view training data, the learned features become more suitable for TDOM/DSM crop registration, leading to improved accuracy and recall.

\noindent\textbf{Ablation on Reference Acquisition Cost.}
We compare the reference-map acquisition cost between the 3D mesh rendering pipeline and our TDOM/DSM crop-based pipeline on UAVD4L-2yr.
The 3D mesh are reconstructed from real-world oblique photogrammetry, which requires oblique photogrammetry, large storage, and online GPU rendering, whereas our method uses compact 2.5D maps and direct cropping.
As shown in Tab.~\ref{tab:ablation_map_cost}, TDOM/DSM reduces the map storage from 19.5~GB to 472~MB, removes the GPU memory requirement during reference acquisition, and achieves faster reference generation.

\noindent\textbf{Comparison with Geo-referenced 3D Mesh References.}
Tab.~\ref{tab:ablation_high_quality_map} further compares our method with PiLoT using geo-referenced 3D mesh references on UAVD4L-2yr.
Despite using only lightweight TDOM/DSM crops, our method achieves comparable or better localization accuracy.
This shows that, with cross-view feature learning and sensor priors, lightweight orthographic maps can serve as an effective alternative to heavy 3D mesh references for UAV geo-localization.

\section{Conclusion}
\label{sec:conclusion}
In this paper, we presented \textbf{PiLoT v2}, a lightweight UAV geo-localization framework that replaces heavy 3D mesh references and online rendering with efficient TDOM/DSM map crops.
By combining cross-view feature registration, multi-hypothesis coarse-to-fine LM optimization, and built-in UAV sensor priors, PiLoT v2 achieves robust pose refinement from lightweight orthographic references.
We also constructed a TDOM/DSM benchmark aligned with existing 3D-mesh-based UAV localization datasets.
Experiments on synthetic, real-world, and long-term sequences show that PiLoT v2 achieves accurate and real-time localization, outperforming existing orthographic-map-based baselines and remaining competitive with 3D-mesh-based localizers.
We believe this work advances vision-based UAV localization under GNSS-denied/degraded conditions and provides useful insights for lightweight map-based localization on other robotic platforms.

\bibliographystyle{IEEEtran}
\bibliography{example}

\end{document}

\maketitle
\thispagestyle{empty}
\pagestyle{empty}

\section{Demo Video}
The supplementary material contains a demonstration video, which shows how PiLoT v2 replaces online 3D rendering with lightweight TDOM/DSM cropping, and achieves robust real-time UAV sequence geo-localization.

%%%%%%%%%%%%%%%%%%%%%%%%%%%%%%%%%%%%%%%%%%%%%%%%%%%%%%%%%%%%%%%%%%%%%%%%%%%%%%%%
\section{Method Details}
\label{sec:supp_method}

\subsection{Network Architecture}
\label{sec:supp_network}

The feature extractor follows a shared two-branch design for the query image and the TDOM reference crop.
Both inputs are processed by the same MobileOne-UNet~\cite{vasu2023mobileone}, so the learned representation is shared across oblique UAV views and orthographic map views.
The encoder is based on the first three stages of MobileOne-S0 with ImageNet-pretrained weights, and the decoder uses three lightweight upsampling blocks with channel widths $[128,64,32]$.
Each decoder output is followed by a descriptor head and a confidence head.
The descriptor head produces a 32-dimensional dense feature map, while the confidence head predicts a single-channel confidence map used to weight feature-metric residuals.
The descriptor maps are L2-normalized along the channel dimension before residual computation.
The detailed configuration of the shared feature extractor is summarized in Tab.~\ref{tab:supp_feature_extractor}.

The output of the network is a three-level feature-confidence pyramid:
\begin{equation}
    \{(\mathbf{F}_{\ell}, \mathbf{W}_{\ell})\}_{\ell=0}^{2},
\end{equation}
where $\mathbf{F}_{\ell}$ denotes the dense descriptor map used to build feature-metric residuals, and $\mathbf{W}_{\ell}$ denotes the corresponding confidence map.

The query image and the reference image share the same feature extractor, implemented by concatenating them along the batch dimension and forwarding them through the same backbone.
During online localization, the reference image is the TDOM crop.
The DSM is not used as an image-like network input; it is only used to lift reference pixels into metric geo-anchors for geometric pose refinement.

\subsection{TDOM/DSM Reference Crop}
\label{sec:supp_reference}
\begin{figure}[t]
    \vspace*{8pt}
    \centering
    \includegraphics[width=\columnwidth]{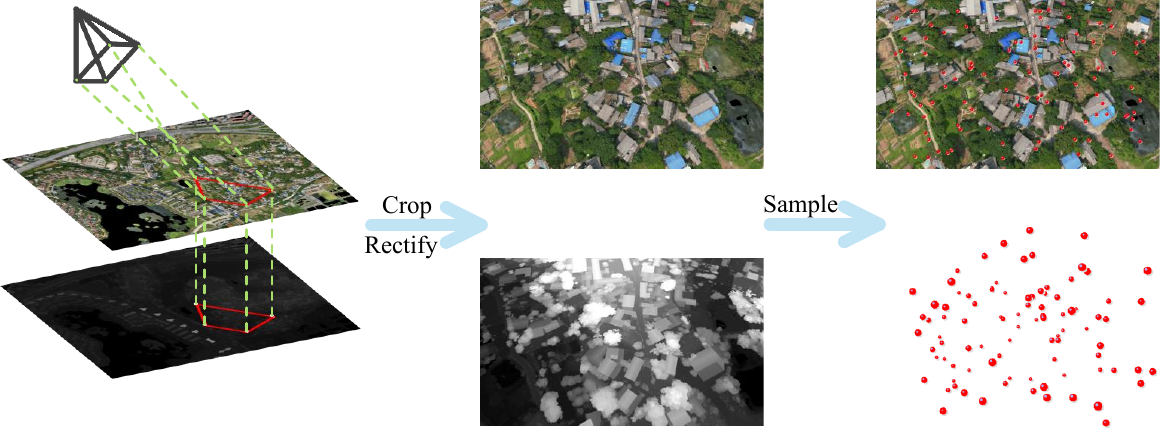}
    \caption{
    \textbf{Illustration of TDOM/DSM reference cropping.}
    The corresponding TDOM, DSM, and map-coordinate grids are cropped and warped into a pixel-aligned 2.5D reference for pose refinement.
    }
    \label{fig:supp_crop_sample}
\end{figure}

The goal of the reference-crop module is to convert the global TDOM/DSM maps into a local pixel-aligned 2.5D reference for the current query frame.
Given the propagated pose prior and the camera intrinsics, the crop is generated through the following three steps, as illustrated in Fig.~\ref{fig:supp_crop_sample}.

\noindent\textbf{Map-footprint Estimation.}
We first estimate the local map footprint covered by the current camera frustum.
For each query frame, rays are emitted from the camera center through the four image corners.
Along each ray, we search for the point that is most consistent with the DSM surface.
Specifically, the selected ray sample is defined as

\begin{equation}
    s_i^\star =
    \arg\min_s
    \left|
    z_{\mathrm{ray}}(s) -
    h_{\mathrm{DSM}}(x_{\mathrm{ray}}(s),y_{\mathrm{ray}}(s))
    \right|.
\end{equation}

The four selected surface points define a quadrilateral footprint on the map.

\begin{table}[t]
\vspace*{8pt}
\centering
\scriptsize
\setlength{\tabcolsep}{3pt}
\renewcommand{\arraystretch}{1.08}
\caption{
Configuration of the shared feature extractor.
}
% 中文表题：共享特征提取器的配置。
\label{tab:supp_feature_extractor}
\resizebox{\columnwidth}{!}{
\begin{tabular}{l l l}
\toprule
Component & Configuration & Role \\
\midrule
Input images
& RGB query / RGB TDOM crop
& Cross-view feature extraction \\
Encoder
& MobileOne-S0, first three stages
& Lightweight multi-scale encoding \\
Pretraining
& ImageNet
& Stable initialization \\
Decoder
& U-Net style, channels $[128,64,32]$
& Multi-scale feature recovery \\
Descriptor head
& $3\times3$ convolution, 32 channels
& Dense feature residuals \\
Confidence head
& $1\times1$ convolution + sigmoid
& Residual weighting \\
Feature normalization
& Channel-wise L2 normalization
& Stabilized feature comparison \\
DSM usage
& Not an image input
& Geo-anchor construction only \\
\bottomrule
\end{tabular}
}
\end{table}

\noindent\textbf{Pixel-aligned Reference Warping.}
The TDOM image, DSM height map, and map-coordinate grids are then warped from the quadrilateral footprint to a rectangular crop using the same homography.
Using the same warp for all map layers ensures that the resulting crop is pixel-aligned across appearance, height, and horizontal map coordinates.
We denote the resulting local 2.5D reference as $(\mathbf{I}^{r}_{i}, \mathbf{D}^{r}_{i})$, where $\mathbf{I}^{r}_{i}$ is the TDOM crop and $\mathbf{D}^{r}_{i}$ is the corresponding DSM crop.

\noindent\textbf{Geo-anchor Construction.}
For each valid reference pixel $(u,v)$, the warped map-coordinate grids provide the horizontal map coordinates $(x_i^{m}(u,v),y_i^{m}(u,v))$.
Together with the DSM height value $\mathbf{D}^{r}_{i}(u,v)$, the pixel is lifted into a metric geo-anchor:

\begin{equation}
    \mathbf{P}^{m}_{i}(u,v)
    =
    [x_i^{m}(u,v),y_i^{m}(u,v),\mathbf{D}^{r}_{i}(u,v)]^\top .
\end{equation}

Valid geo-anchors are sampled from the reference crop and used in the subsequent feature-metric pose refinement.
In this way, the crop module replaces online 3D rendering with a lightweight map-warping operation while still providing metric 3D anchors for 6-DoF optimization.

\subsection{Multi-Hypothesis Refinement}
\label{sec:supp_multihyp}

\begin{figure*}[t]
    \vspace*{8pt}
    \centering
    \includegraphics[width=\textwidth]{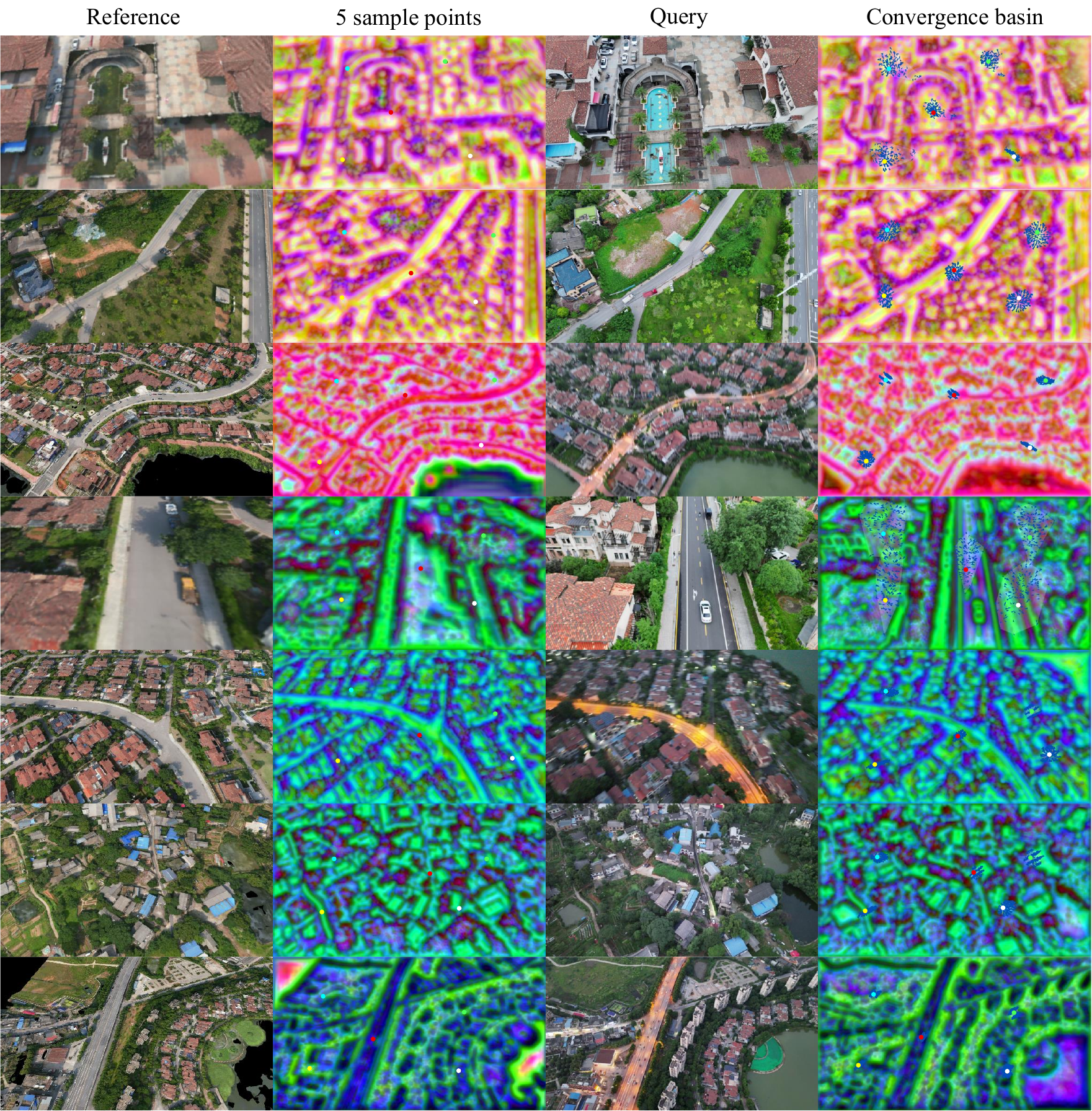}
    \caption{
    \textbf{Visualization of TDOM-to-query refinement.}
    Reference geo-anchors sampled from the TDOM/DSM crop are projected into the query image under different pose hypotheses.
    The arrow field illustrates the update direction induced by the learned cross-view feature-metric residuals, showing how candidate projections are guided toward consistent query locations.
    }
    \label{fig:supp_convergence}
\end{figure*}

The optimizer uses a multi-hypothesis initialization to improve robustness against pose-prior errors, as visualized in Fig.~\ref{fig:supp_convergence}.
In the current implementation, horizontal translation hypotheses are sampled on a 2D grid in the local map plane with a range of $\pm 10$ m and a step size of 2 m along both $x$ and $y$ directions.
Yaw hypotheses are sampled with the offsets

\begin{equation}
    \Delta\psi \in
    \{\pm1^\circ,\pm3^\circ,\pm5^\circ,\pm7^\circ,\pm9^\circ\}.
\end{equation}

The Cartesian product of the horizontal translation and yaw candidates gives 1210 initial hypotheses.
After the coarse-level optimization, the hypotheses are ranked by the post-update objective value, and only the top 128 candidates are retained for the following finer levels.
This strategy enlarges the convergence basin when the propagated pose prior has horizontal translation or yaw errors.

\subsection{Sensor-Prior Fusion in LM}
\label{sec:supp_sensor_lm}

The testing system enables both the single-point laser range prior and the gravity direction prior during LM refinement.
The single-point laser range prior is implemented as a height-consistency residual between the center-ray range point and the DSM surface.
Given the measured range, the center image ray is back-projected from the query camera to obtain a 3D point.
The residual is computed as the height difference between this point and the DSM height at the corresponding horizontal map location.
This term anchors the pose to the local DSM surface and reduces the depth ambiguity of the 2.5D reference.

The gravity prior is implemented as a gravity-direction consistency residual.
Given a candidate pose, its rotation predicts the gravity direction in the camera frame, which is compared with the IMU-measured gravity direction.
This prior constrains the camera tilt with respect to the vertical direction, mainly roll and pitch, while yaw is not directly constrained by gravity.

The visual, range, and gravity terms are fused in the same LM normal equation.
In implementation, their Hessian and gradient contributions are accumulated as

\begin{equation}
    \mathbf{H}_{\mathrm{total}}
    =
    \mathbf{H}_{\mathrm{vis}}
    +
    \alpha_{\mathrm{range}} \mathbf{H}_{\mathrm{range}}
    +
    \alpha_{\mathrm{grav}} \mathbf{H}_{\mathrm{grav}},
\end{equation}

\begin{equation}
    \mathbf{g}_{\mathrm{total}}
    =
    \mathbf{g}_{\mathrm{vis}}
    +
    \alpha_{\mathrm{range}} \mathbf{g}_{\mathrm{range}}
    +
    \alpha_{\mathrm{grav}} \mathbf{g}_{\mathrm{grav}}.
\end{equation}

The scales $\alpha_{\mathrm{range}}$ and $\alpha_{\mathrm{grav}}$ are adjusted according to the relative gradient magnitudes and are further controlled by residual-dependent gates.
When a sensor-prior residual becomes excessively large, its contribution is down-weighted to avoid corrupting the visual registration.

%%%%%%%%%%%%%%%%%%%%%%%%%%%%%%%%%%%%%%%%%%%%%%%%%%%%%%%%%%%%%%%%%%%%%%%%%%%%%%%%
\section{Cross-view Training Details}
\label{sec:train}

This section provides the training details that are not expanded in the main paper.
We focus on the data sources, training-pair curation, sample construction, and unrolled geometric supervision used for learning the cross-view feature representation.

\subsection{Training Data Sources and Curation}
\label{sec:supp_training_data}

The training data contains two complementary sources, as summarized in Tab.~\ref{tab:supp_training_data}.
The cross-view training data provides oblique-to-orthographic pairs, which adapt the model to TDOM-style orthographic references used during online localization.
The original PiLoT training set~\cite{cheng2026pilot} provides oblique-to-oblique pairs, which preserve the geometry-aware registration ability learned from perspective-view image pairs.

\begin{table}[t]
\vspace*{8pt}
\centering
\scriptsize
\setlength{\tabcolsep}{3pt}
\renewcommand{\arraystretch}{1.08}
\caption{
Statistics of the training data sources.
}
\label{tab:supp_training_data}
\resizebox{\columnwidth}{!}{
\begin{tabular}{l l c c}
\toprule
Source & Pair Type & Sequences & Raw Pairs \\
\midrule
Cross-view training data
& Oblique-to-orthographic
& 375
& 1,125,000 \\
Original PiLoT training set
& Oblique-to-oblique
& 82
& 1,268,351 \\
\bottomrule
\end{tabular}
}
\end{table}

The verified oblique-to-orthographic pairs are curated before training to reduce sampling bias.
For this source, the maximum usage count of each orthographic reference image is limited to 5, preventing a small number of map references from dominating the training distribution.
During each epoch, these verified oblique-to-orthographic pairs are mixed with 50,000 oblique-to-oblique pairs sampled from the original PiLoT training source.
This mixed training strategy biases the network toward the orthographic-reference setting while retaining geometric supervision from perspective-view pairs.

\subsection{Training Sample Construction}
\label{sec:supp_training_input}

Each training sample consists of one RGB query image, one RGB reference image, camera intrinsics and poses, and a precomputed per-query 3D point cloud.
Both images are resized by the shorter side and then cropped to $512 \times 512$.
The network input contains only RGB images; TDOM, DSM, and depth maps are not used as image-like inputs during training.
Depth maps are only used offline to generate the precomputed 3D point clouds and geometric annotations.

For each training sample, 500 reference geo-anchors are sampled to compute the geometric reprojection supervision.
The initial query pose is generated by perturbing the ground-truth query pose, while the reference pose remains fixed to the ground truth.
The translation perturbation is sampled from 5 to 15 meters, and the rotation perturbation is sampled from 5 to 15 degrees.

\subsection{Unrolled Reprojection Supervision}
\label{sec:supp_training_objective}

Training is performed by unrolling the feature-metric pose refinement process.
Starting from the perturbed query pose, the optimizer repeatedly projects the sampled 3D points into the query feature map, constructs feature-metric residuals against the reference features, and solves a LM-style update for the 6-DoF pose increment.
The feature residuals drive the pose update, while the final supervision is applied on the geometric reprojection error after each refinement step.

Let $\mathbf{T}^{\star}_{q}$ denote the ground-truth query pose and $\mathbf{T}^{k}_{q}$ denote the query pose after the $k$-th refinement step.
For a sampled 3D point $\mathbf{P}_{j}$, the ground-truth projection and the optimized projection are
\begin{equation}
    \mathbf{p}^{\star}_{j}
    =
    \pi(\mathbf{K}^{q}, \mathbf{T}^{\star}_{q}\mathbf{P}_{j}),
    \qquad
    \mathbf{p}^{k}_{j}
    =
    \pi(\mathbf{K}^{q}, \mathbf{T}^{k}_{q}\mathbf{P}_{j}).
\end{equation}
The per-step reprojection loss is defined as
\begin{equation}
    \mathcal{L}_{k}
    =
    \frac{1}{|\mathcal{V}|}
    \sum_{j\in\mathcal{V}}
    \rho_{B}
    \left(
    \left\|
    \mathbf{p}^{k}_{j}
    -
    \mathbf{p}^{\star}_{j}
    \right\|_{2}^{2}
    \right),
\end{equation}
where $\mathcal{V}$ denotes the visible point set and $\rho_{B}(\cdot)$ is the Barron robust loss~\cite{barron2019robustloss}.
In our implementation, the refinement is unrolled over nine coarse-to-fine steps, and the final training objective averages the reprojection losses over all unrolled steps.

\subsection{Training Schedule}
\label{sec:supp_training_schedule}

The model is trained for 30 epochs using Adam with an initial learning rate of $1\times10^{-3}$ and zero weight decay.
The learning rate is decayed by a StepLR scheduler every 10 epochs with a decay factor of 0.25.
We use 8 RTX 4090 GPUs with a batch size of 12 per GPU.

%%%%%%%%%%%%%%%%%%%%%%%%%%%%%%%%%%%%%%%%%%%%%%%%%%%%%%%%%%%%%%%%%%%%%%%%%%%%%%%%
\section{Runtime Analysis}
\label{sec:supp_runtime}

PiLoT v2 follows a sequential localization setting, where the local map reference is updated continuously while the incoming UAV video stream is processed frame by frame.
A purely serial implementation would stall pose estimation until the TDOM/DSM reference of the current frame is generated.
To reduce this waiting time, the online system is organized into two asynchronous threads.
The map-preparation thread generates the local TDOM/DSM reference crop for subsequent frames, while the pose-localization thread estimates the 6-DoF pose of the current query frame.
The offline TDOM/DSM map construction cost is excluded from all online runtime measurements.

\subsection{Online runtime breakdown}
The per-module runtime is summarized in Tab.~\ref{tab:supp_runtime}.
The TDOM/DSM crop is executed in the map-preparation thread and is reported separately because it runs asynchronously with the localization thread.
The pose-localization thread consists of geo-anchor back-projection, dense feature extraction, pose optimization, and post-processing.
Among these components, the optimizer is the dominant part of the online localization time.

\begin{table}[t]
\vspace*{8pt}
\centering
\scriptsize
\setlength{\tabcolsep}{3pt}
\renewcommand{\arraystretch}{1.08}
\caption{
Online runtime breakdown of PiLoT v2.
}
\label{tab:supp_runtime}
\resizebox{\columnwidth}{!}{
\begin{tabular}{l l c}
\toprule
Module & Description & Time (ms) \\
\midrule
\multicolumn{3}{l}{\textit{Map-preparation thread}} \\
TDOM/DSM crop
& Local reference generation
& 13.17 \\
\midrule
\multicolumn{3}{l}{\textit{Pose-localization thread}} \\
Back-projection
& Geo-anchor conversion and pose initialization
& 3.84 \\
Feature extraction
& Shared dense feature pyramid
& 8.44 \\
Optimizer
& Coarse-to-fine LM refinement
& 43.82 \\
Post-processing
& Candidate filtering and pose conversion
& 2.67 \\
\midrule
Total pose-localization time
& --
& 60.51 \\
\bottomrule
\end{tabular}
}
\end{table}
\begin{table}[t]
\vspace*{8pt}
\centering
\scriptsize
\setlength{\tabcolsep}{3pt}
\renewcommand{\arraystretch}{1.08}
\caption{
Runtime attribution inside the pose optimizer.
}
\label{tab:supp_opt_runtime}
\resizebox{\columnwidth}{!}{
\begin{tabular}{l l c}
\toprule
Group & Component & Time (ms) \\
\midrule
Optimizer
& Residual/Jacobian evaluation
& 29.45 \\

Optimizer
& Cost-only objective evaluation
& 7.12 \\

Optimizer
& Batched LM solve
& 2.22 \\

Optimizer
& Other overhead
& 5.03 \\
\midrule
Residual evaluation
& Fused visual residual
& 1.32 \\
Residual evaluation
& Range / gravity priors
& 17.88 \\
Residual evaluation
& Prior setup and transfer
& 1.21 \\
Residual evaluation
& Visual--sensor system assembly
& 8.58 \\
Residual evaluation
& Other overhead
& 0.46 \\
\midrule
Total optimizer time
& --
& 43.82 \\
\bottomrule
\end{tabular}
}
\end{table}

\subsection{Optimizer implementation and runtime attribution}
The pose optimizer contains two computationally different parts: regular feature-metric operations and irregular map/sensor-prior operations.
The feature-metric term is evaluated over many geo-anchor--hypothesis pairs and has a regular memory-access pattern on dense feature maps.
We therefore implement its system construction with a fused CUDA kernel.
For each geo-anchor under a pose hypothesis, the kernel performs projection, visibility checking, descriptor interpolation, feature-residual evaluation, local feature-gradient estimation, confidence weighting, and accumulation of the corresponding gradient and Hessian contribution.
This design keeps the most repetitive visual computations inside a compact GPU execution path, avoiding explicit intermediate tensors and reducing kernel-launch overhead.

The range prior requires pose-dependent DSM surface queries along the center ray, and the gravity prior requires candidate-wise attitude conversion and Jacobian evaluation.
These operations involve coordinate transformations, irregular DSM access, finite-difference computation, and CPU--GPU synchronization.
They provide important physical constraints for stabilizing 6-DoF refinement with TDOM/DSM references, but they cannot be fused into the same image-grid kernel used by the visual term.
Consequently, the runtime is no longer dominated by the fused visual residual itself; a large fraction of the optimizer cost is spent on sensor-prior evaluation and the assembly of the combined visual--sensor optimization system.

The optimizer runtime is attributed in Tab.~\ref{tab:supp_opt_runtime}.
The batched LM solve accounts for only a small portion of the optimizer time, since each hypothesis only requires solving a small damped $6\times6$ system on the GPU.
Most of the runtime is spent before the solve, where the optimizer repeatedly evaluates residuals, forms Jacobians, fuses sensor-prior contributions, and the updated hypotheses are then re-evaluated by a cost-only objective pass for candidate ranking.
The regular visual registration term is efficiently accelerated by CUDA fusion, while the physically grounded range and gravity priors introduce irregular map-access and coordinate-conversion costs.

%%%%%%%%%%%%%%%%%%%%%%%%%%%%%%%%%%%%%%%%%%%%%%%%%%%%%%%%%%%%%%%%%%%%%%%%%%%%%%%%

\section{Experimental Results}

\subsection{Weather Condition Robustness}
\label{sec:supp_weather}

To further analyze robustness to appearance variations, we regroup the SynthCity-6 test sequences according to weather and illumination conditions.
Specifically, the same SynthCity-6 evaluation set used in the main experiments is divided into six subsets: sunny, cloudy, sunset, rainy, foggy, and night.
Tab.~\ref{tab:supp_weather_robustness} reports the localization recall of each method.

The results show that PiLoT v2 maintains stable recall across different weather and illumination conditions. 
Compared with the TDOM/DSM crop-based baselines, PiLoT v2 achieves consistently stronger recall at strict and medium thresholds, indicating that the learned cross-view feature registration and sensor-aided pose refinement improve robustness under appearance variations.
Compared with 3D render-based methods, PiLoT v2 remains competitive while using lightweight TDOM/DSM references instead of online mesh rendering. In several conditions, PiLoT or other render-based baselines achieve higher R@1, because mesh-based methods generate viewpoint-dependent reference images from full 3D scene geometry, which can better match the perspective appearance of the query image. However, PiLoT v2 shows strong stability at medium and loose thresholds. The advantage is especially clear under rainy and night conditions, where appearance degradation makes pure visual matching more difficult.
The learned feature registration, multi-hypothesis initialization, and sensor-aided LM refinement help guide the estimate back to a reliable pose.

These results further support the robustness of the proposed crop-based localization framework on SynthCity-6 under diverse visual conditions.
\begin{table}[t]
\vspace*{8pt}
\centering
\scriptsize
\setlength{\tabcolsep}{3pt}
\renewcommand{\arraystretch}{1.08}
\caption{
Pitch-range robustness.
Each cell reports Recall@1/3/5/10 (m,$^\circ$) (\%).
}
\label{tab:supp_pitch_robustness}

\resizebox{\columnwidth}{!}{
\begin{tabular}{lcc}
\toprule
\multirow{2}{*}{Method}
& $[-90^\circ,-70^\circ]$
& $[-90^\circ,-40^\circ]$ \\
&
Recall@1/3/5/10
& Recall@1/3/5/10 \\
\midrule

Render2Loc-eLoFTR~\cite{yan2023rendercompare}
& 19.57 / 78.01 / 93.02 / 99.40
& 66.67 / 92.92 / 97.95 / 99.75 \\

Render2Loc-RoMa~\cite{yan2023rendercompare}
& \textbf{67.24} / 96.51 / 99.36 / 99.98
& 77.76 / 96.26 / 99.22 / 99.96 \\

PixLoc~\cite{sarlin2021pixloc}
& 54.47 / 89.53 / 89.98 / \textbf{100.00}
& 78.33 / 96.19 / 97.33 / 98.72 \\

PiLoT~\cite{cheng2026pilot}
& 61.23 / \textbf{99.53} / \textbf{99.99} / \textbf{100.00}
& 82.81 / 98.12 / 98.17 / 98.27 \\

OrthoLoC-eLoFTR\cite{dhaouadi2026ortholoc}
& 28.70 / 92.40 / 99.70 / \textbf{100.00}
& 77.46 / 99.71 / \textbf{100.00} / \textbf{100.00} \\

OrthoLoC-RoMa\cite{dhaouadi2026ortholoc}
& 31.20 / 90.69 / 98.99 / 99.99
& 75.51 / 99.49 / \textbf{100.00} / \textbf{100.00} \\

\rowcolor{ourcolor}
\textbf{PiLoT v2 (Ours)}
& 49.15 / 99.21 / 99.98 / \textbf{100.00}
& \textbf{85.17} / \textbf{99.99} / \textbf{100.00} / \textbf{100.00} \\

\bottomrule
\end{tabular}
}
\end{table}

\begin{table}[t]
\vspace*{8pt}
\centering
\scriptsize
\setlength{\tabcolsep}{3pt}
\renewcommand{\arraystretch}{1.08}
\caption{
Effect of multi-hypothesis sampling.
}
\label{tab:supp_seed_ablation}

\resizebox{\columnwidth}{!}{
\begin{tabular}{lcccc}
\toprule
Sampling
& Med.m$\downarrow$
& Med.$^\circ\downarrow$
& R@1/3/5/10$\uparrow$ 
& FRC$\downarrow$ \\
\midrule

None
& 2.363
& 1.662
& 4.75 / 63.72 / 79.22 / 90.08 
& 28 \\

$\psi$ only
& 1.917
& 1.504
& 5.97 / 75.67 / 93.28 / 96.92 
& 3 \\

$(x,y)$ only
& 1.752
& 1.445
& 6.44 / 80.39 / 92.95 / 98.36 
& 8 \\

\rowcolor{ourcolor}
\textbf{PiLoT v2 (Ours)}
& \textbf{1.666}
& \textbf{1.398}
& \textbf{6.58} / \textbf{87.08} / \textbf{98.92} / \textbf{100.00}
& \textbf{0} \\

\bottomrule
\end{tabular}
}
\end{table}

\begin{table*}[t]
\vspace*{8pt}
\centering
\scriptsize
\setlength{\tabcolsep}{3pt}
\renewcommand{\arraystretch}{1.08}
\caption{
Weather-condition robustness.
Each cell reports Recall@1/3/5/10 (m,$^\circ$) (\%).
}
\label{tab:supp_weather_robustness}

\resizebox{\textwidth}{!}{
\begin{tabular}{l c c c c c c}
\toprule
\multirow{2}{*}{Method}
& Sunny
& Cloudy
& Sunset
& Rainy
& Foggy
& Night \\
&
Recall@1/3/5/10
& Recall@1/3/5/10
& Recall@1/3/5/10
& Recall@1/3/5/10
& Recall@1/3/5/10
& Recall@1/3/5/10 \\
\midrule

Render2Loc-eLoFTR
& 49.38 / 88.87 / 96.82 / 99.78
& 48.49 / 87.00 / 95.95 / 99.69
& 48.84 / 88.29 / 96.82 / 99.82
& 65.26 / 91.48 / 97.11 / 99.45
& 44.97 / 87.72 / 96.36 / 99.72
& 38.28 / 80.19 / 93.17 / 99.09 \\

Render2Loc-RoMa
& 61.44 / 93.61 / 98.74 / 99.86
& 60.97 / 93.33 / 98.62 / 99.84
& 61.01 / 93.36 / 98.64 / 99.85
& 62.98 / 93.05 / 98.57 / 99.83
& 69.60 / 96.34 / 98.88 / 99.80
& 61.03 / 92.54 / 98.56 / 99.73 \\

PixLoc
& 64.82 / 79.49 / 79.60 / 99.82
& 77.31 / 99.67 / 99.78 / 99.91
& 78.24 / \textbf{100.00} / \textbf{100.00} / \textbf{100.00}
& 78.19 / 98.85 / 99.04 / 99.44
& 68.29 / 94.98 / 96.38 / 98.76
& 55.33 / 91.91 / 95.18 / 97.22 \\

PiLoT
& \textbf{74.20} / 94.51 / 94.53 / 94.62
& \textbf{77.75} / 99.62 / 99.73 / 99.84
& \textbf{78.45} / \textbf{100.00} / \textbf{100.00} / \textbf{100.00}
& 78.04 / 99.67 / 99.81 / \textbf{100.00}
& \textbf{73.97} / \textbf{99.92} / \textbf{100.00} / \textbf{100.00}
& 63.80 / 98.97 / \textbf{99.97} / \textbf{100.00} \\

OrthoLoC-eLoFTR
& 58.73 / 98.89 / 99.91 / \textbf{100.00}
& 57.38 / 97.09 / 99.78 / \textbf{100.00}
& 58.53 / 96.91 / 99.91 / \textbf{100.00}
& 76.52 / 99.71 / \textbf{100.00} / \textbf{100.00}
& 54.64 / 96.45 / 99.95 / \textbf{100.00}
& 50.83 / 94.92 / 99.69 / \textbf{100.00} \\

OrthoLoC-RoMa
& 58.24 / 95.98 / 99.47 / 99.98
& 57.44 / 95.62 / 99.49 / \textbf{100.00}
& 59.06 / 96.04 / 99.47 / \textbf{100.00}
& 77.44 / 99.52 / \textbf{100.00} / \textbf{100.00}
& 50.20 / 94.80 / 99.75 / \textbf{100.00}
& 53.31 / 95.75 / 99.50 / \textbf{100.00} \\

\rowcolor{ourcolor}
\textbf{PiLoT v2 (Ours)}
& 72.51 / \textbf{99.80} / \textbf{100.00} / \textbf{100.00}
& 72.05 / \textbf{99.71} / \textbf{100.00} / \textbf{100.00}
& 71.00 / 99.73 / \textbf{100.00} / \textbf{100.00}
& \textbf{82.18} / \textbf{100.00} / \textbf{100.00} / \textbf{100.00}
& 67.92 / 99.72 / 99.97 / \textbf{100.00}
& \textbf{64.53} / \textbf{99.22} / \textbf{99.97} / \textbf{100.00} \\

\bottomrule
\end{tabular}
}
\end{table*}

\subsection{Pitch Range Robustness}
\label{sec:supp_pitch}

In addition to weather and illumination changes, we further analyze the influence of pitch on localization performance.
Here, $-90^\circ$ denotes a nadir-facing camera, which is closest to the orthographic view.
Tab.~\ref{tab:supp_pitch_robustness} evaluates two pitch-range settings: a near-orthographic setting of $[-90^\circ,-70^\circ]$ and a wider setting of $[-90^\circ,-40^\circ]$.

The near-orthographic setting is challenging for most methods.
Although this setting appears closer to the orthographic map view, it also suppresses perspective cues from facades and vertical structures.
So PiLoT v2 improves over the TDOM/DSM crop-based baselines, but it does not outperform all 3D render-based methods at the strictest threshold.
When the pitch range is extended to $[-90^\circ,-40^\circ]$, PiLoT v2 achieves the best recall.
This suggests that the proposed cross-view feature learning can better exploit these cues when aligning the query image with the TDOM/DSM reference.
Compared with OrthoLoC variants, PiLoT v2 is more stable across both pitch settings, indicating that feature learning and multi-sensor refinement are important for handling pitch-induced viewpoint variations.

\subsection{Effect of Multi-hypothesis Sampling}
\label{sec:supp_multi_hypothesis_ablation}

We further evaluate the effect of multi-hypothesis sampling in the pose refinement stage.
Since the TDOM/DSM reference crop is generated from a pose prior, errors in horizontal translation and yaw can lead to an inaccurate local map reference and degrade the subsequent feature-metric registration.
Therefore, we compare four initialization strategies: no additional seed, yaw-only sampling, horizontal translation-only sampling, and joint $(x,y,\psi)$ sampling.

As shown in Tab.~\ref{tab:supp_seed_ablation}, removing the multi-hypothesis seeds leads to clear performance degradation.
Sampling yaw or horizontal translation alone improves the results, indicating that both sources of pose-prior error affect the optimization.
The best overall performance is achieved by joint $(x,y,\psi)$ sampling, which obtains the lowest median errors and the highest recall at all reported thresholds.
This confirms that the proposed multi-hypothesis initialization enlarges the convergence basin of the optimizer and improves robustness against inaccurate pose priors.

\subsection{Qualitative Trajectory Results}
\label{sec:supp_qualitative}

\begin{figure*}[t]
    \vspace*{8pt}
    \centering
    \includegraphics[width=\textwidth]{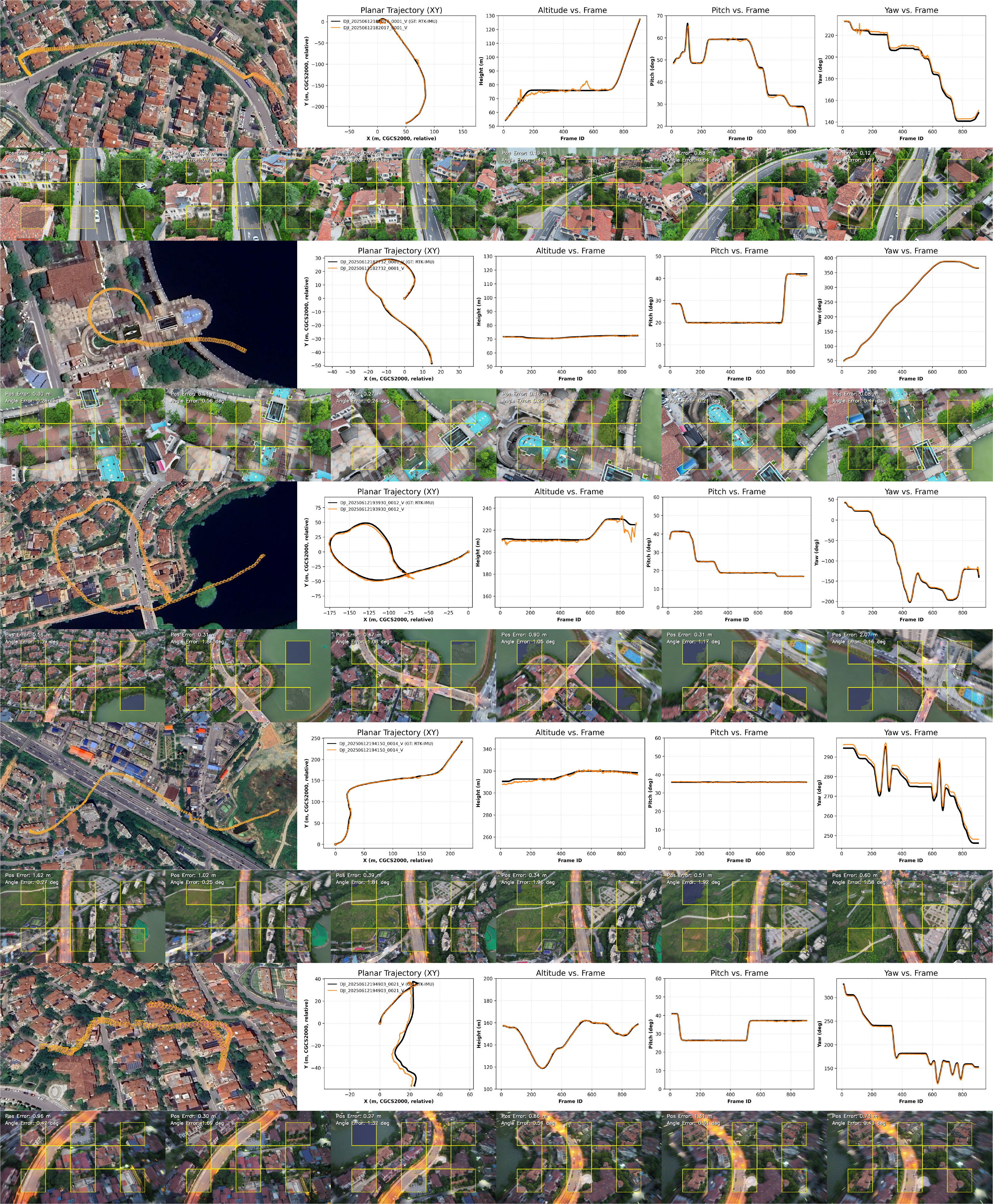}
    \caption{
    \textbf{Trajectory estimation results on challenging real-world and long-term scenarios.}
    This figure demonstrates the model’s robustness by evaluating on 4 real-world sequences from the UAVD4L-2yr dataset, which include challenging day/night conditions.
    }
    \label{fig:feicuiwan}
\end{figure*}

\begin{figure*}[t]
    \centering
    \includegraphics[width=\textwidth]{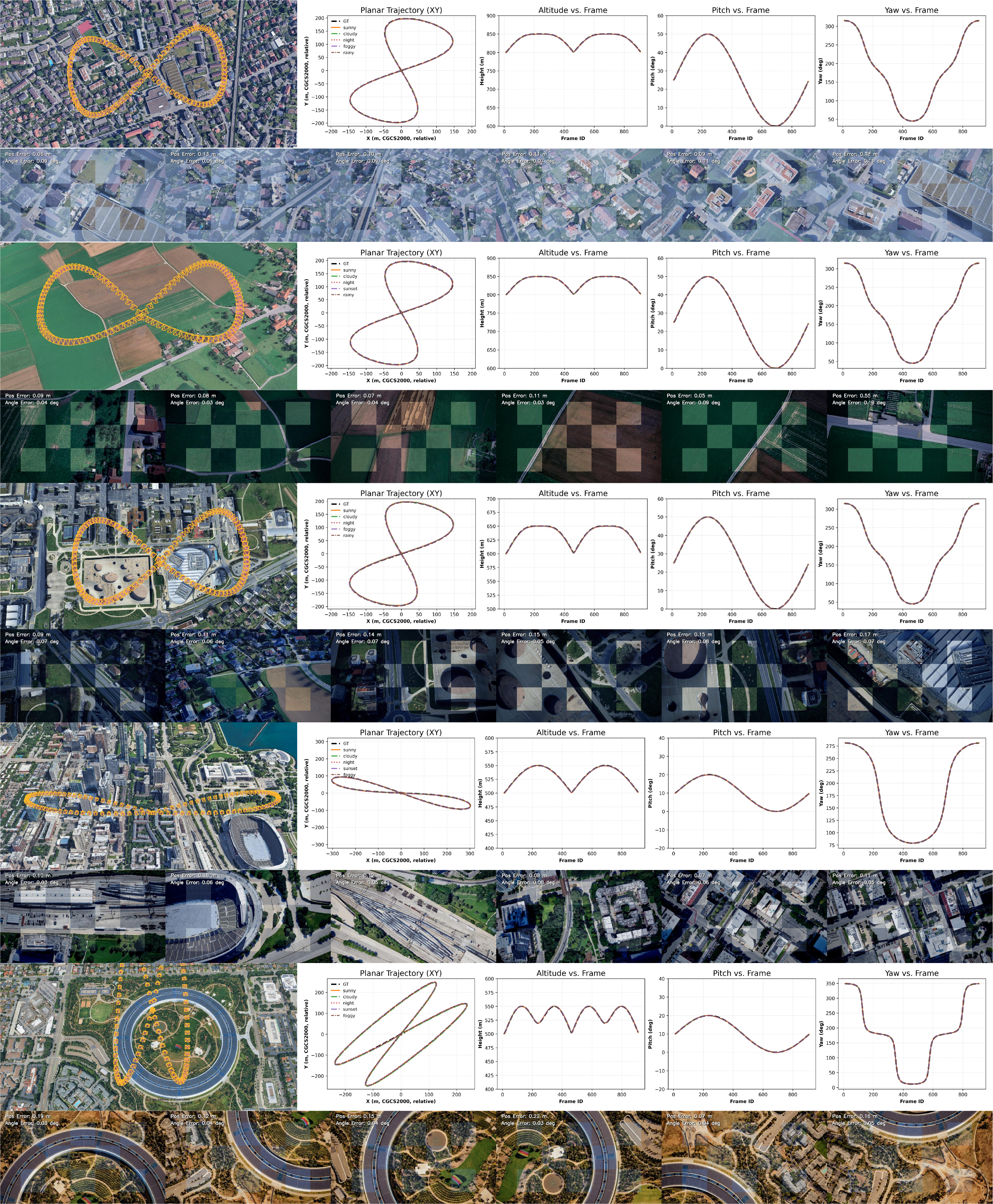}
    \caption{
    \textbf{Trajectory estimation results on various synthetic scenes from the SynthCity-6 dataset.}
    The figure showcases our method’s performance across diverse synthetic conditions, including various scenes and weather.}
    \label{fig:google}
\end{figure*}

\begin{figure*}[t]
    \vspace*{8pt}
    \centering
    \includegraphics[width=\textwidth]{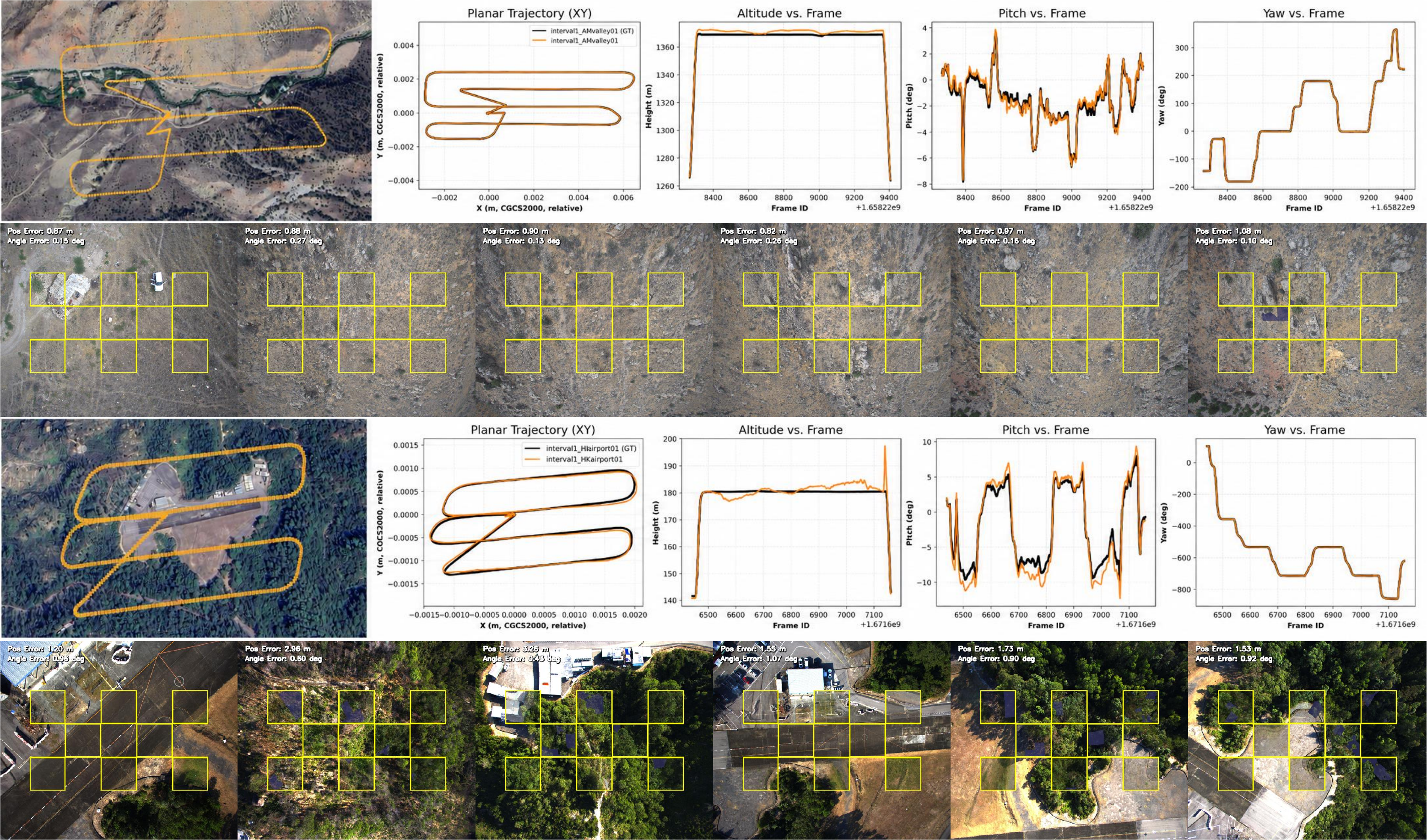}
    \caption{
    \textbf{Generalization performance on the UAVScenes benchmark dataset.}
    This figure validates our model’s strong generalization capability on two diverse, unseen scenes from the public UAVScenes benchmark. 
    }
    \label{fig:uavscene}
\end{figure*}

We further provide qualitative trajectory visualizations on real-world, synthetic, and public benchmark scenes.
Fig.~\ref{fig:feicuiwan} shows results on challenging real-world and long-term UAVD4L-2yr sequences.
Fig.~\ref{fig:google} presents trajectory results on diverse SynthCity-6 synthetic scenes.
Fig.~\ref{fig:uavscene} further reports the generalization performance on the UAVScenes benchmark.

%%%%%%%%%%%%%%%%%%%%%%%%%%%%%%%%%%%%%%%%%%%%%%%%%%%%%%%%%%%%%%%%%%%%%%%%%%%%%%%%

\bibliographystyle{IEEEtran}
\bibliography{example}